\pdfoutput=1
\documentclass[11pt]{article}

\usepackage[final]{acl}

\usepackage{times}
\usepackage{latexsym}

\usepackage[T1]{fontenc}
\usepackage[utf8]{inputenc}

\usepackage{microtype}

\usepackage{inconsolata}

\usepackage{graphicx}

\usepackage{array}
\usepackage{amsmath}
\usepackage{amssymb}
\usepackage{mathtools}
\usepackage{amsthm}
\usepackage{multirow}
\usepackage{booktabs}
\usepackage{xspace}
\usepackage{xcolor}
\usepackage{colortbl}
\title{QiMeng-Attention: SOTA Attention Operator is generated by \\ SOTA Attention Algorithm}

\author{
 \textbf{Qirui Zhou\textsuperscript{1,3}},
 \textbf{Shaohui Peng\textsuperscript{2}},
 \textbf{Weiqiang Xiong\textsuperscript{2,3}},
 \textbf{Haixin Chen\textsuperscript{1,3}},
 \textbf{Yuanbo Wen\textsuperscript{1}},\\
 \textbf{Haochen Li\textsuperscript{2}},
 \textbf{Ling Li\textsuperscript{2,3}\thanks{Corresponding Author.}},
 \textbf{Qi Guo \textsuperscript{1}},
 \textbf{Yongwei Zhao\textsuperscript{1}},
 \textbf{Ke Gao\textsuperscript{2}},
 \textbf{Ruizhi Chen\textsuperscript{2}},\\
 \textbf{Yanjun Wu\textsuperscript{2}},
 \textbf{Chen Zhao\textsuperscript{2}},
 \textbf{Yunji Chen\textsuperscript{1,3}}
\\
 \textsuperscript{1}SKL of Processors, Institute of Computing Technology, CAS, Beijing, China\\
 \textsuperscript{2}Intelligent Software Research Center, Institute of Software, CAS, Beijing China\\
 \textsuperscript{3}University of Chinese Academy of Sciences, Beijing, China
\\
\tt\small  zhouqirui22s@ict.ac.cn,\{pengshaohui,liling\}@iscas.ac.cn 
}
\newcommand{\name}{LLM-TL\xspace}
\usepackage{listings}
\lstdefinestyle{mystyle}{
  basicstyle=\ttfamily\footnotesize,
  breakatwhitespace=false,         
  breaklines=true,                 
  captionpos=b,                    
  keepspaces=true,                 
  numbersep=5pt,                  
  showspaces=false,                
  showstringspaces=false,
  showtabs=false,                  
  tabsize=2,
  frame=single
}
\begin{document}
\maketitle

\begin{abstract}
% 可以去掉
%Unlike traditional intelligence algorithms, Large Language Models (LLMs) enable cross-domain autonomous intelligence.
% attn是最重要的算子（尤其是long seq），bottleneck
% The attention operator, particularly FlashAttention, is crucial for high-performance acceleration in LLMs.
%As the core layer of the Transformer, 
%The attention operator has become a bottleneck for the performance of large language models (LLMs), especially in long-context scenarios.
The attention operator remains a critical performance bottleneck in large language models (LLMs), particularly for long-context scenarios. 
% The attention operator, particularly FlashAttention, is crucial for high-performance acceleration in LLMs. 
%现在有很多针对复杂GPu特性的加速算法，（例如flash attn），必须需要人实现，time consuming。
% FlashAttention is the most commonly used method for accelerating the attention operator in consideration of GPU characteristics, but due to the heterogeneity and complexity of different GPUs, they all require time-consuming and inefficient manual implementation.
While FlashAttention is the most widely used and effective GPU-aware acceleration algorithm, it must require time-consuming and hardware-specific manual implementation, limiting adaptability across GPU architectures.
%However, the increasing heterogeneity of GPUs challenges the efficient implementation of FlashAttention algorithms. Manual implementations are inefficient and non-portable, while existing compilers cannot fully optimize the operator. 
%现在LLM能生成代码，但对于复杂的attn还不行，只能生成high-level的，且效率不高，
% 挑战：不能理解GPU特性和利用优化技术，以生成高性能attn kernel。
% Recent LLMs have shown a lot of promise in code generation tasks, but they only can generate un-optimized high-level language code for attention operator with limited performance and efficiency. 
% %attention implementation using low-level GPU characteristics.
% %The key challenge is LLMs can hardly understand the low-level GPU characteristics and complexity of attento generate implementi attention optimization.
% % face challenges creating optimized code on the first try. This makes it necessary to use other strategies at inference time to generate optimized code. 
% % The primary challenge lies in the incapability of LLMs to comprehend the intricately complex attention operator and determine how to perform targeted optimizations by harnessing the underlying characteristics of GPUs.
% The key challenge is that LLMs struggle to understand the complex data flow and computation process of the attention operator on GPUs, thus failing to optimize it by leveraging GPU memory hierarchy and Tensor core.
Existing LLMs have shown a lot of promise in code generation tasks, but struggle to generate high-performance attention code.
The key challenge is it cannot comprehend the complex data flow and computation process of the attention operator and utilize low-level primitive to exploit GPU performance.
%optimization logic of of GPU memory hierarchies and Tensor Core operations.

% Thinking language是为了让大模型理解融合，
To address the above challenge, we propose an LLM-friendly Thinking Language (\name) 
%by decoupling optimization logic from low-level implementation,
to help LLMs decouple the generation of high-level optimization logic and low-level implementation on GPU, and enhance LLMs' understanding of attention operator.
% so that can understand and optimize attention operator more easily.
%大模型分别理解和生成语义级别优化和底层kernel实现.
% \name consists of dataflow and implementation statements for GPU optimizations.
Along with a 2-stage reasoning workflow, TL-Code generation and translation, the LLMs can automatically generate FlashAttention implementation on diverse GPUs, establishing a self-optimizing paradigm for generating high-performance attention operators in attention-centric algorithms.
Verified on A100, RTX8000, and T4 GPUs, the performance of our methods significantly outshines that of vanilla LLMs, achieving a speed-up of up to $35.16\times$.
Besides, our method not only surpasses human-optimized libraries (cuDNN and official library) in most scenarios but also extends support to unsupported hardware and data types, reducing development time from months to minutes compared with human experts.
% Our methodand also outperforms cuDNN and official implementation in most cases.
% %Moreover, the development time also be reduced up to $1,248\times$ compared with the senior Coder.
% Moreover, the development time can also be reduced from months to minutes compared with human experts.
%Experiment results.
\end{abstract}

\section{Introduction}

% 对于attention的描述太泛了，访存和计算怎么复杂
% Compared to traditional intelligence algorithms restricted by specific domains and labeled data, autonomous intelligence based on Large Language Models (LLMs) extends across diverse fields and applications.
% The attention operator, a core computation in LLMs, has inspired numerous high-performance acceleration algorithms.
% FlashAttention is particularly prominent among them.
% However, the continuous evolution in the heterogeneity and complexity of GPUs, the dominant intelligent computing hardware, has made the efficient implementation and migration of FlashAttention algorithms an urgent challenge.
%As the core layer within the Transformer architecture, the attention operator 
The attention mechanism is the cornerstone of modern Large Language Models (LLMs). Its time and memory complexity grows quadratically in sequence length, and attention has emerged as the bottleneck for the runtime and memory resource requirements for LLMs, particularly in long-context scenarios \cite{dao2022flashattention}.
% has emerged as a bottleneck for the runtime and memory resources requirement for Large Language Models (LLMs), and further dictates the performance of LLMs, particularly in long-context scenarios \cite{dao2022flashattention}.
%Unlike traditional intelligence algorithms limited by specific domains and labeled data, autonomous intelligence driven by Large Language Models (LLMs) has shown cross-domain applicability and ~\cite{guo2023evaluating}.
%The attention operator, core computation in LLMs, has led to many high-performance acceleration algorithms, with FlashAttention~\cite{dao2022flashattention} being particularly notable.
Consequently, numerous acceleration algorithms have been proposed to accelerate attention operators,
%for high-performance attention operator implementation on GPU have emerged,
with FlashAttention being the most effective and widely used \cite{dao2022flashattention,dao2024flashattention-2,hong2024flashdecodingfasterlargelanguage}.
With the rise of cost-efficient, high-performance LLMs like DeepSeek~\cite{deepseekv2,deepseekv3,deepseekr1}, deploying LLMs across diverse GPU specifications (including legacy generations) has become increasingly prevalent.

\begin{figure}[!t]
    \centering
    \includegraphics[width=1.0\linewidth]{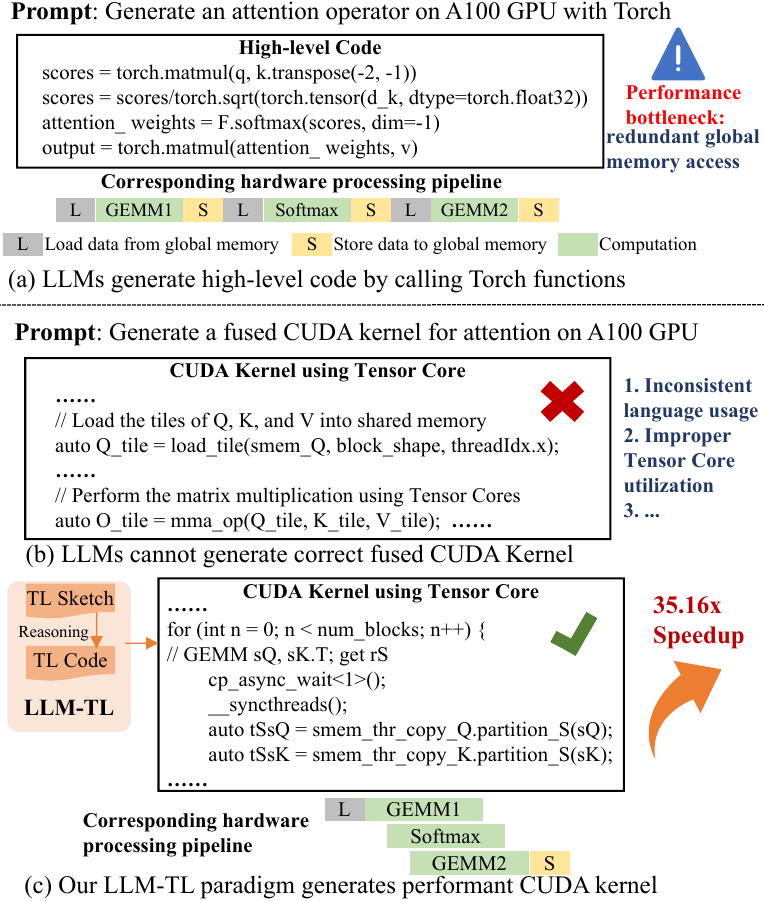}
    \caption{Our \name enables LLMs to generate high-performance Tensor Core kernel for attention.}
    % \caption{Comparison between generation paradigm through LLM-friendly Language (LLM-TL) and other methods.}
    \label{fig:intro}
\end{figure}

% Currently, the FlashAttention operator is mainly implemented manually.
% Human experts perform fusion optimization in attention operators based on a deep understanding of GPU characteristics, effectively enhancing the computation efficiency of LLMs.
% However, considering the complexity of the fusion implementation and the differences in parameters and instruction set architectures among different GPUs, this manual implementation is not only time-consuming but also hardly portable, thus fails to fully support commonly used GPUs and data types (such as the RTX8000 platform and FP8 data type).
% Existing deep learning compilers (such as TVM and Ansor) cannot perform fusion optimization for the FlashAttention operator, thus failing to fully utilize the GPU's performance.
% In general, existing methods cannot automatically understand and utilize GPU characteristics to efficiently generate high-performance FlashAttention operators on different GPU platforms.

However, due to the complexity of attention operators and disparities in the architectures and instruction sets across different GPUs, the efficient implementation and migration of FlashAttention algorithms becomes a pressing challenge.
%However, due to the complexity of attention operators and disparities in the architectures and instruction sets across different GPUs, 
%the FlashAttention operator is mainly implemented manually.
Existing methods for implementing high-performance attention operators can be divided into two categories: i.e., manual implementation and deep learning compilers. 
%There are two main paradigms for implementing high-performance operators, i.e., manual implementation and deep learning compilers. 
Human experts optimize attention operators through fused computation based on their expertise in GPU architecture and algorithm knowledge, thereby effectively enhancing computation efficiency.
%However, due to the complexity of the attention operator and disparities in the architectures and instruction sets across different GPUs,
Consequently, these hand-crafted implementations are time-consuming and non-portable, failing to support some common GPUs and datatypes (e.g., RTX8000 GPU and FP8 datatype).
% In addition, existing common deep learning compilers (e.g., TVM~\cite{chen2018tvm}, Ansor~\cite{zheng2020ansor}) fail to automatically implement the FlashAttention operator, because the leading to limited exploitation of the GPU's performance.
In addition, existing mainstream deep learning compilers (e.g., TVM~\cite{chen2018tvm}, Ansor~\cite{zheng2020ansor}) fail to automatically implement FlashAttention due to their inability to perform optimizations involving fused sequences of multiple complex operations (e.g. GEMM).
% Overall, current methods struggle to automatically use GPU characteristics for efficient high-performance FlashAttention operator generation on different GPUs.
Overall, current methods struggle to automatically analyze the attention operator characteristics and GPU architecture to efficiently generate high-performance FlashAttention implementation on different GPU platforms, which limits the adaptability of LLMs.
% The remarkable capabilities of LLMs in the field of code generation offer new possibilities to address the above challenges, and thus are expected to achieve a self-optimizing paradigm for automatically generating high-performance attention operators in attention-centered inference.
% Currently, although LLMs can generate simple layer-by-layer implementations of attention operators by using deep learning frameworks (e.g., PyTorch), they struggle to generate high-performance FlashAttention operators based on low-level primitives (like CUDA or CuTe) to fully unleash the GPU's computing power.
% The core reason lies in the huge gap between the semantic knowledge of LLMs and the fine-grained optimizations on GPUs.
% LLMs cannot understand and apply GPU characteristics through reasoning in high-level languages (e.g., natural language or Python) to achieve high-performance optimization of FlashAttention operators, such as data blocking and reuse, as well as fused computation.
% LLMs' strong code-generation abilities~\cite{bairi2024codeplan, zhong2024can,holtl2mac,Li2024CONSIDERCA, wang2023codet5opencodelarge, Nijkamp2022CodeGenAO} offer new ways to tackle these issues and may achieve a self-optimizing paradigm for generating high-performance attention operators in attention-centered algorithms.

Motivated by strong code-generation abilities of LLMs~\cite{bairi2024codeplan, zhong2024can,holtl2mac,Li2024CONSIDERCA, wang2023codet5opencodelarge, Zhou_Wen_Chen_Gao_Xiong_Li_Guo_Wu_Chen_2025, zhang2025qimengtensoropautomaticallygeneratinghighperformance}, it is a promising direction to tackle these issues by utilizing LLMs automatically implement FlashAttention on GPUs.
% as a self-optimizing paradigm for generating high-performance attention operators in attention-centered algorithms.
Although LLMs can create attention implementations by high-level APIs like PyTorch~\cite{pytorch}, they have trouble generating high-performance FlashAttention with low-level primitives (e.g., CUDA or CuTe~\cite{cute}) with crucial optimization techniques like data blocking and fused computation.
This significantly hinders the reduction of memory access overheads and the enhancement of computational efficiency, as shown in Figure \ref{fig:intro}(a).%due to the semantic gap between their knowledge and GPU characteristics.
% They can't use high - level language reasoning to optimize FlashAttention operators for high performance.
The primary challenges preventing LLMs from generating high-performance FlashAttention implementations in one pass are twofold:
(1) LLMs cannot comprehend the complex data flow and multi-step computation of attention, thus failing to generate effective optimization logic;
(2) LLMs are unable to produce optimized implementations based on low-level primitives to fully exploit GPU characteristics.

Inspired by human experts' decoupling of optimization and implementation, we propose an LLM-friendly Thinking Language (\name) to address the above challenge by helping LLMs decouple the generation of abstract optimization logic and low-level code implementation on GPU.
Specifically, \name encompasses two types of statements and their requisite parameters: memory access and computation, enabling LLMs to comprehend the data flow and computation processes of the attention operator on GPU architecture from the vantage point of high-level abstract semantics, as shown in Figure \ref{fig:intro}(c).
Based on \name, we further propose an automated workflow, leveraging the reasoning capabilities of LLMs to implement the high-performance FlashAttention operator on GPUs with low-level primitives.% in a self-optimizing manner.
Formally, the workflow includes: 1) TL Code generation: LLMs use LLM-TL statements to create sketch code to represent the execution flow on GPU, then reasoning and filling the parameters needed for each statement; 
2) TL Code translation: LLMs convert the TL-code into low-level CUDA code based on the target GPU architecture and instruction set.
Through \name and the workflow, LLMs can efficiently and automatically implement the high-performance FlashAttention on different GPUs.

% Besides, we also propose an automated workflow to enable LLMs to leverage \name to implement the high-performance FlashAttention operator on GPUs efficiently, thus achieving the self-optimizing paradigm.
%\name includes dataflow statements for abstract GPU optimizations and implementation statements for specific details.
% The workflow includes: 1) TL-code generation: LLMs use LLM-TL primitives to create sketch code to represent the execution flow on attention on GPU, then reasoning and filling the parameters needed for each primitive; 
%2) TL-Code generation: Parameter expressions are inferred from input shapes and GPU parameters; 
% 2) Adaptive translation: LLMs convert the TL-code into low-level CUDA code based on the target GPU architecture and instruction set.
% Through \name and the workflow, LLMs can efficiently and automatically implement the high-performance FlashAttention on different GPUs.

The contributions of this paper are as follows:
 
    (1) We propose \name, an LLM-friendly abstract language, enabling LLMs to deeply comprehend and optimize complex attention operators on GPUs, thereby establishing a self-optimizing paradigm for generating high-performance attention operators in attention-centric algorithms.
    
    (2) We present an automated workflow that enables LLMs to leverage \name to implement high-performance FlashAttention on diverse GPUs, which extends the capabilities of LLMs.
    
    (3) Verified on A100, RTX8000, and T4 GPUs, the performance of FlashAttention implementation generated by our methods significantly outshines that of vanilla large language models (LLMs), achieving a speed-up of up to $35.16\times$, and also outperforms with cuDNN or official implementation in most cases.
Moreover, the development time also be reduced from months to minutes compared with human experts.

\section{Preliminary}

% GPU Characteristics
%% memory hierarchy
%% tensor core and mma
% attention operator and acceleration
%% MHA,GQA,MQA,and MLA
%% FlashAttention acceleration

%preliminary部分是否有对FlashAttention加速的现有工作的介绍？
\subsection{GPU Characteristics}

\begin{figure}
    \centering
    \includegraphics[width=0.8\linewidth]{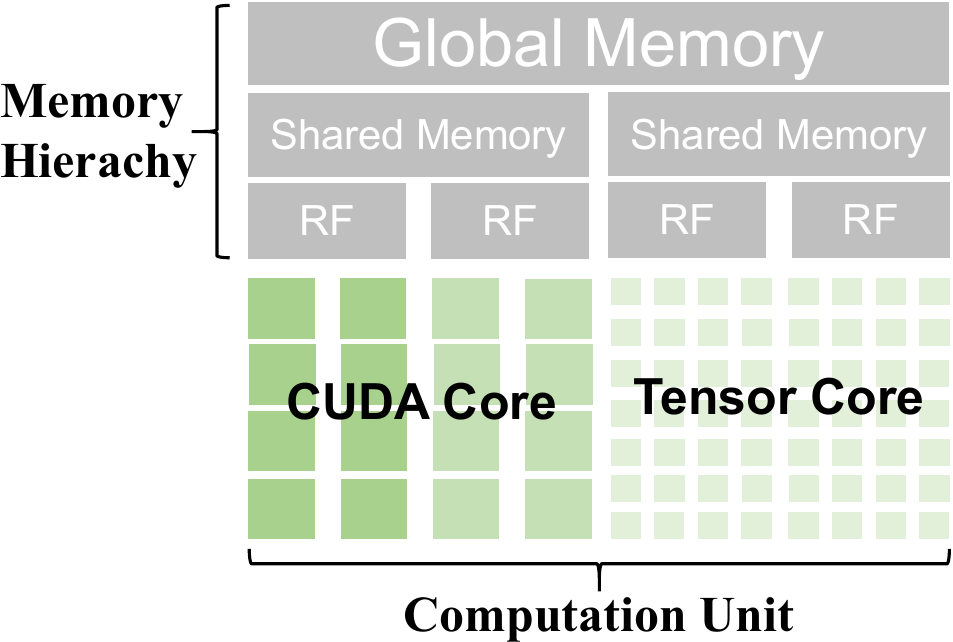}
    \caption{Demonstration of GPU architecture.
    %GPU programming model. The programming model of GPU (CUDA) is structured into three levels: grid, thread block, and thread, each corresponding to a distinct hierarchy of memory reuse.
    }
    \label{fig:GPU}
    \vspace{-10pt}
\end{figure}

The rapid development of AI is closely tied to the advancement of GPU's high parallel computing power. GPUs achieve high parallelism by stacking a large number of simple computation units. They use the SIMT (Single Instruction, Multiple Threads) design, which allows for the execution of many threads running the same code. The introduction to the GPU architecture is illustrated in Figure \ref{fig:GPU}.

\subsubsection{Memory Hierarchy}
% The memory hierarchy of GPUs is primarily divided into three levels: Memory, Cache, and Register. Among these, the external memory of a GPU is commonly referred to as Global Memory. To address the locality requirements of data access by different threads, a high-speed cache controllable by programmers, known as Shared Memory, is introduced in GPU programming for storing data shared among threads. This design results in a unique memory model for GPUs: all threads share Global Memory, threads within the same thread block (Block) share Shared Memory, while each thread has its own private Registers. Due to this complex and hierarchical memory structure, programming requires explicit specification of the size, location, and source and destination memory levels for data transfers between different layers. For data stored in registers, it can be directly accessed for fast computations, which typically include and General Matrix Multiplication (GEMM) and other fundamental operations, with results stored back in registers for subsequent use. 
The memory hierarchy of GPUs is divided into global memory, shared memory, and registers. Global memory is shared by all threads, shared memory is shared by threads within the same thread block, and each thread has its own private registers. Due to this hierarchical memory structure, explicit management of data movement between different memory levels is required during programming. Data stored in registers can be directly accessed for fast computations, such as General Matrix Multiplication (GEMM).

\subsubsection{Computation Unit}
To get better performance on GPU, we use Tensor Core~\cite{tensorcore}, which is a specialized processing unit optimized for matrix operations. Unlike standard CUDA Cores, Tensor Cores use the warp (a cluster of 32 threads) as the basic computing unit, enabling efficient register data utilization and coordination among threads within a warp. Due to the design characteristics of Tensor Cores, operators can more fully utilize register resources within threads, which provides better hardware support for operator fusion optimization.
%Unlike standard CUDA Core, Tensor Core utilizes the warp as the fundamental computing unit, fully leveraging data sharing and coordination among threads within a warp, thereby significantly accelerating matrix operations on GPUs. Furthermore, its unique register reuse mechanism further reduces memory access latency, providing hardware-level acceleration support for operator fusion.
% Starting with the Volta architecture, NVIDIA introduced Tensor Core~\cite{tensorcore}, specialized computing units optimized for matrix operations. Unlike standard CUDA Cores, Tensor Cores use the warp as the basic computing unit, enabling efficient data utilization and coordination among threads within a warp. This allows 32 threads to process larger matrix blocks in parallel within a single instruction and output complete matrix multiplication results, significantly accelerating matrix operations on GPUs. Additionally, due to the design characteristics of Tensor Cores, operators can more fully utilize register resources within threads, improving data reuse rates in higher-level memory, which provides better hardware support for operator fusion optimization.

The efficient utilization of Tensor Core typically requires direct manipulation of the CUDA PTX instruction set, a low-level programming approach closer to assembly language, involving numerous complex index calculations.
To simplify such process, NVIDIA introduced CuTe~\cite{cute}, an advanced template librariy which encapsulate the underlying PTX instructions through a high-level abstraction layer and automate index calculations. Compared to PTX, CuTe provides a more user-friendly and efficient programming support for automatic generation.

Based on these characteristics of GPUs, we abstract the execution flow of operators into two basic types of operations: data movement statement (describing the transfer of data between different memory levels) and computation statement (describing computing operations), thereby clearly delineating the execution process of operators on GPUs.

\subsection{Attention Operators and acceleration}
The Attention mechanism computes relevance weights among input elements to focus on key information for effective feature extraction and integration.
The input consists of sequences $Q, K, V $ of dimension $d$ and number of tokens $N$. The attention mechanism computes the output using the formula $\text{Attention}(Q,K,V)=\text{Softmax}(\text{Mask}(\frac{QK^T}{\sqrt d}))V$. 
In addition to the early Multi-Head Attention (MHA)~\cite{attention}, recent advancements have introduced Multi-Query Attention (MQA)~\cite{shazeer2019mqa} and Group-Query Attention (GQA)~\cite{ainslie2023gqa}. Furthermore, DeepSeek recently proposed a new attention variant called Multi-Head Latent Attention (MLA)~\cite{deepseekv2}. 
These attention variants use fewer key/value heads or employ low-rank joint compression for keys and values. As a result, they significantly reduce the KV cache size, and are therefore widely adopted in contemporary LLMs.
%These advanced attention operators are all designed to reduce the volume of $K,V$ data so that minimize the overhead associated with KV cache, which has led to their widespread adoption in contemporary LLM applications.
%By compressing and reconstructing input embeddings, MLA further reduces the storage overhead of KV cache without losing information. 
% These innovative attention operators not only demonstrate the ongoing optimization of attention mechanisms in academia but also provide crucial support for efficient model training and inference.
% These attention operators 

% As LLMs become increasingly widespread and their computational demands grow, the performance requirements for their core operator—the attention mechanism—are also rising. A prominent example of optimizing the attention operator on GPUs is FlashAttention. FlashAttention fuses two GEMM operations, dividing the matrix multiplications of Q, K, and V into blocks and accumulating them, allowing each block of Q, K, and V to be loaded and stored only once while completing the computation of two matrix multiplications on-chip. For the intermediate Softmax layer, FlashAttention employs the online softmax algorithm, enabling global softmax to be computed iteratively in blocks. This avoids the need to write the entire matrix back to global memory and reload it into cache for computation, significantly reducing spatial complexity and efficiently utilizing high-level storage variables for fused computation.

\begin{figure*}[]
    \centering
    \includegraphics[width=0.9\linewidth]{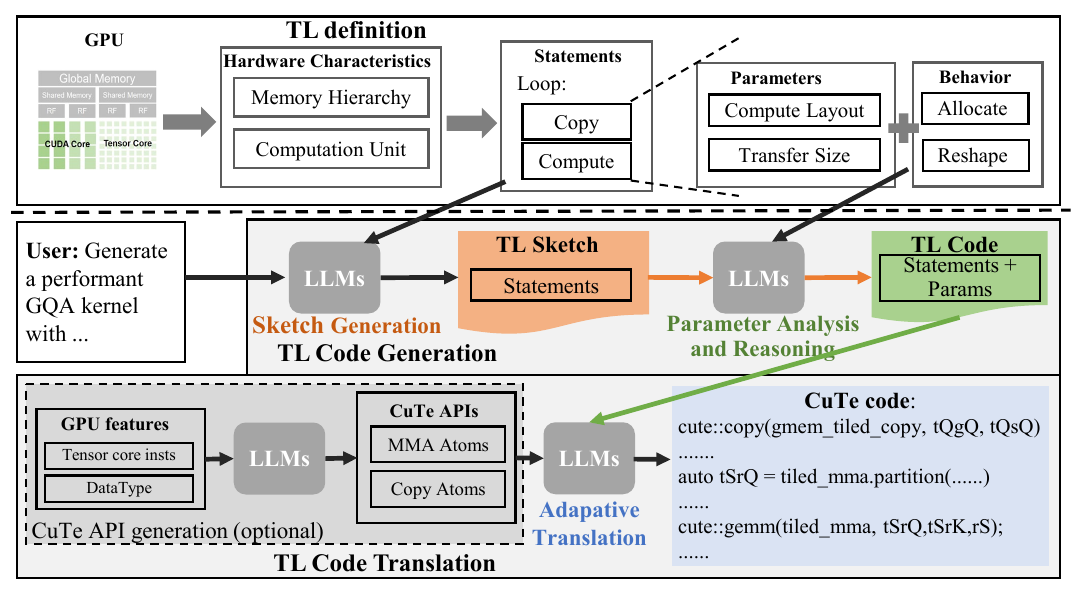}
    \caption{\textbf{\name overview.} We design the Thinking Language (TL) to help LLMs describe attention execution workflows and parameters on GPUs. Our approach consists of two stages, TL Code Generation and Translation. The TL Sketch, representing abstract semantic execution flow, is initially generated from user requirements. Subsequently, the LLM infers the parameter details of the statements within the Sketch. Finally, TL code will be translated to the CuTe implementation by LLMs based on GPU specification.}
    \label{fig:method}
    \vspace{-10pt}
\end{figure*}

To enhance the efficiency of attention operators, numerous optimization methods have emerged. FlashAttention ~\cite{dao2022flashattention, dao2024flashattention-2} is a prominent example of optimizing the attention operator on GPUs. 
%It utilizes employs the online softmax algorithm, and fuses two GEMM operations, divides the matrix multiplications of $Q$, $K$ and $V$ into blocks and accumulates them, allowing each block to be loaded and stored only once while completing two matrix multiplications on-chip. 
%Meanwhile, it employs the online softmax algorithm, enabling global softmax to be computed iteratively in blocks, avoiding the need to write the entire matrix back to global memory, significantly reducing spatial complexity, and efficiently utilizing high-level storage variables for fused computation.
% High-performance attention operators require deeply integrating the algorithm with GPU hardware characteristics. 
%复杂度高，优化难度高，Tensor Core难
However, the implementation of high-performance attention operators like FlashAttention is often highly complex, requiring extensive low-level optimizations, precise computational scheduling, and facing significant challenges in operator fusion. Additionally, the programming model for GPU hardware, particularly Tensor Cores, is inherently complex, demanding developers to have a deep understanding of its architectural characteristics. As a result, the development and application of attention operators still encounter substantial technical challenges.
%However, the rapid proliferation of novel attention operators, coupled with the inherent complexity of GPU architectures—particularly Tensor Cores—and the intricate design of high-performance attention mechanisms such as FlashAttention, has rendered the generation of efficient attention operators on GPUs a formidable challenge. The intricate process of attention fusion on GPUs further exacerbates this difficulty, necessitating sophisticated optimization strategies to harness the full potential of hardware capabilities while maintaining computational efficiency.
%However, implementing the FlashAttention operator requires a thorough understanding of GPU hardware architecture, Tensor Core programming, and algorithmic workflows. Additionally, as attention variants continue to proliferate, development complexity further increases, leading to a significant rise in development costs.

\section{Method}

%方法部分第一段，是整个方法的Overview，介绍方法一共有哪些部分组成，简述每个部分通过什么实现了什么，最后介绍整个Workflow是什么结构。
%示例：In this section, we propose an LLM-friendly Thinking Language (\name).
% By decoupling the generation of abstract optimization and low-level code implementation on GPU, \name enable the LLM to comprehend the data flow and computation processes of the attention operator on GPU architecture, thereby reasoning high-performance optimization code from the vantage point of high-level abstract semantics.
%As shown in Figure~\ref{}, \name encompasses two types of primitives: memory access and computation, and it delineates the requisite parameters for these primitives.
% Based on \name, we propose an end-to-end workflow, errnabling LLMs to implement the high-performance FlashAttention operator on GPUs efficiently, thus achieving the self-optimizing paradigm.
%Specifically, after obtaining the primitives, we apply a TL-code generation step to ..., then we further introduce a Adaptive translation step to .... .
% 方法部分名字的层级好像对不上。目前图2里面，最高层级的名字有两个：TL Code Generation和TL Code Translation。低层级的有3个：Generate process description， Analysis and reasoning和Adaptive Translation。最好层级能统一。

%考虑到大模型无法直接生成GPU上复杂的底层FlashAttention实现（以CuTe为例，通常包含数百行CUDA代码）。因此我们引入了模糊了实现细节，仅保留抽象语义逻辑的LLM-TL表示（仅需十几行便可表示上述实现），并驱动大模型先生成TL code，在将TL code逐步翻译为底层代码实现的方式完成该任务。
%这一节我们将分别介绍TL code的生成和翻译过程。
%To obscure implementation details and retain only semantic-level optimization logics, \name can condense the aforementioned implementation into just over ten lines of semantic code.
\name solely incorporates statements that delineate the processing flow of attention and the essential parameters required, thereby enabling the reduction of hundreds of lines of low-level CUDA code to a mere dozen lines of TL code.
This simplification empowers Large Language Models (LLMs) to comprehend and implement high-performance optimizations of attention operators on GPUs, encompassing fusion and computational tensorization.
We introduce the workflow of TL code generation and adaptive translation from TL code to lower-level CuTe code in the following subsections, as shown in Figure \ref{fig:method}.
%first, driving the LLM to generate TL code, and then progressively translating it into low-level code implementation. This section will elaborate on the processes of TL code generation and translation.
%The components and workflow are as shown in Figure \ref{fig:method}.

\subsection{LLM-friendly Thinking Language (LLM-TL) Definition}
To assist LLMs in generating code more effectively, we abstract the execution process on GPUs into two types of statements: \textit{Copy} for memory access and \textit{Compute} for various types of computations. These statements are used to represent the execution flow and parameter details of attention operators on GPUs. 
By summarizing GPU characteristics into semantic-level descriptions and hiding unnecessary details, TL can be better leveraged through the generalization and semantic understanding capabilities of LLMs.

\subsection{TL Code Generation}
Specifically, the generation of TL Code is divided into two steps: first, describe the execution flow of attention into a TL Sketch containing computation and memory access statements, and then analysis and reason to supplement detailed statement parameters, ultimately producing a complete TL Code.
\subsubsection{Sketch Generation}
To facilitate the gradual understanding of GPU hardware architecture and attention operator characteristics by LLMs, we first drive LLMs to generate the execution flow of the attention operator into a semantically structured representation, TL Sketch, which consists of two fundamental statements: \textit{Compute} and \textit{Copy} hierarchies according to the algorithm's execution flow. Due to the various attention operator variants and GPUs, it is necessary to abstract high-level semantics of the attention execution on general GPU architecture to make the LLMs generate optimization logic. Thus, inspired by concepts from computer architecture, we design efficient statements, copy and compute, for two purposes: comprehensively capture the semantic-level optimization logic (multi-level memory data movement and GPU computation units utilization) and effectively hide low-level implementation details.

Specifically, for \textit{Copy} statements, LLMs use \textit{Copy} to describe the movement of tensors between different memory hierarchies (global memory, shared memory, and registers).
For example, the clause \textit{Copy Q from global to shared} indicates loading Q from global memory into shared memory.
For \textit{Compute} statements, LLMs can represent necessary internal computations (such as GEMM, arithmetic operations, etc.) on the corresponding tensors.
Based on TL statements, LLM can incorporate GPU hardware features to optimize attention execution at a semantical level, like representing the fusion optimization by continuously arranging multiple \textit{Compute} at the same memory hierarchy without any \textit{Copy}.

\subsubsection{Parameter Analysis and Reasoning}
TL Sketch can describe the execution flow of the attention operator on GPUs at a semantic level, but it lacks the details information needed to generate executable code.
For \textit{Copy} statement, it requires address allocation behavior and parameters like block dim to specify the transfer size;
while for \textit{compute}, it needs parameters to specify the computation layout and sometimes conduct reshape operations to connect adjacent computations to achieve fusion.
%ude the shape information of intermediate variables to clarify the computational specifics.

Based on the above summary, we leverage the reasoning capabilities of LLMs to supplement the parameter details required for \textit{Copy} and \textit{Compute} statements in the TL Sketch, such as block sizes and shape transformation information, to obtain the final TL-code.
Taking the FlashAttention operator as an example, each thread block loads a fixed batch and head of \textit{Q} with dimensions \textit{(BM,HeadDim)}, where \textit{BM} represents the block size of the query sequence length, and \textit{HeadDim} denotes the dimensionality of a single head in multi-head attention (MHA). At this point, the \textit{Copy} statement for Q is expanded to \textit{Copy Q (BM, HeadDim) in coor [L = block\_idx] from global to shared memory}, where \textit{block\_idx} denotes the block index in CUDA programme, adding the location and size information of Q in global memory. To fuse two consecutive GEMM operations at the register level, a reshape statement is required to perform shape inference for both GEMM operations. Then a reshape statement is introduced, specifically \textit{reshape rS from mma\_C to mma\_A}\footnote{Tensor Cores use \textit{mma} (matrix multiply-accumulate) instructions to achieve , where matrices A, B, and C need to follow hardware-defined layouts. Here the \textit{mma\_A} and \textit{mma\_C} represent the corresponding layout of each tensor tile.}, to facilitate the transformation and alignment of data structures between the matrix multiplication operations. 
% 把其他的行为（allocate，reshape）也加到这里

% Additionally, to more comprehensively describe the dimension information of implicit variables (such as input/output variables and some intermediate variables), we introduce the allocate statement and allow the LLM to infer the variables and dimension information that need to be explicitly defined based on the semantics of the Sketch. This approach enables us to generate TL Code containing complete algorithm execution information, making it easier for the LLMs to produce accurate code.

\subsection{TL Code Translation}

\subsubsection{CuTe Framework}
We employ the CuTe framework to implement the final GPU code, aiming to simultaneously balance the high performance of GPU computing and the reliability of code generation by LLMs.
The CuTe framework significantly enhances the semantic expressiveness of the code by providing high-level abstractions over fundamental Tensor Core operations (such as the matrix multiply-accumulate instruction \textit{mma} and the matrix load instruction \textit{ldmatrix}).
%说下atom和tensor core指令的关系
In CuTe, assembly-level PTX \textit{mma} instructions are encapsulated into corresponding classes, which include parameters such as computation scale and data types, while also reducing the need for index calculations.
This strategy endows CuTe with a richer semantic hierarchy compared to native Tensor Core instruction programming and can fully leverage the strengths of LLMs in semantic understanding and logical reasoning. For some of the newer architectures, CuTe may not include the corresponding MMA APIs. However, due to their relatively fixed structure, we can quite easily prompt the LLM to generate the corresponding MMA through few-shot learning.
%, thereby generating GPU code that is both correct and highly efficient. Through this approach, we achieve a harmonious integration of the executability of LLM-generated code and the computational performance of GPUs.

\begin{figure}[]
    \centering
    \includegraphics[width=0.9\linewidth]{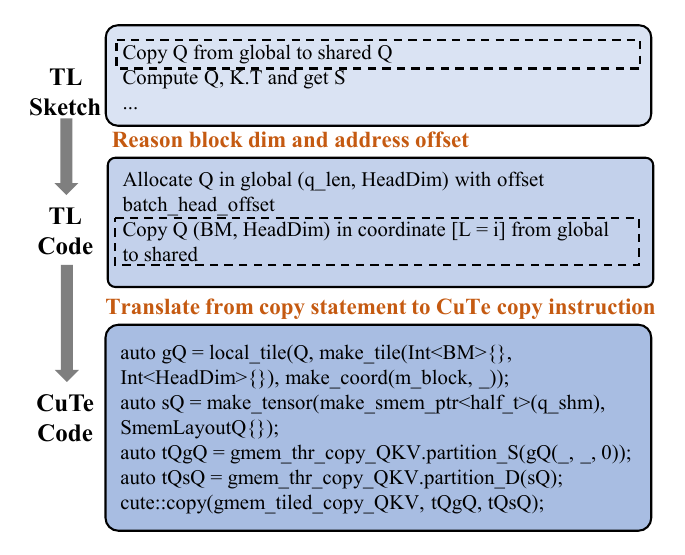}
    \caption{The content within the dashed box illustrates the generation process from a \textit{Copy} statement in TL Sketch to CuTe Code.}%, including parameter reasoning for a \textit{Copy} statement and translation to CuTe APIs.} %Here shows how to reason the TL Code and then translate it into the CuTe implementation for a Copy statement.}
    \label{fig:enter-label}
    \vspace{-15pt}
\end{figure}

\vspace{-5pt}
\subsubsection{Adaptive Translation}
Based on the TL Code, which contains comprehensive attention operator execution details, each statement can be accurately translated into the corresponding CuTe code.Thanks to the decoupling feature of our TL design, each statement can be fully and precisely translated into executable CuTe code, ensuring a smooth transition from high-level specifications to low-level implementations. 
Taking the TL Code as input, we provide the necessary execution information, such as CuTe MMA Atom and \textit{Copy} Atom, for the specific hardware architecture in the prompt. By combining these two parts, we can achieve the translation of TL Code across multiple platforms.
For instance, an allocate operation before \textit{Copy} in the TL Code can be implemented using CuTe's APIs for tensor definitions across various memory hierarchies (such as global memory, shared memory, and registers), in conjunction with the parameter information provided in the TL Code.
Similarly, a \textit{Copy} statement can be realized through CuTe's definitions of source and destination memory (including caches or registers), along with the \textit{cute::copy} API and other related code blocks.
Figure \ref{fig:enter-label} shows some details of how to generate the final CuTe code for a \textit{Copy} statement.

% Building on the aforementioned characteristics, we have designed a universal prompt template to guide the LLM in generating executable code for GPUs across different architectures. Specifically, after obtaining the TL Code with complete semantic and parameter information, we design hardware-agnostic prompts for each statement in the TL Code to facilitate its translation into CuTe code. Due to variations in instruction types across GPU architectures of different generations (e.g., the Tensor Core instruction scale is m16n8k8 for Turing architecture GPUs, while it is m16n8k16 for Ampere architecture), we provide the LLM with the necessary instruction information for the current GPU architecture to generate CuTe code tailored to the specific GPU. Notably, for certain newer architectures or data types, CuTe may not yet natively support the corresponding mma instructions. However, thanks to the regularity of CuTe's instruction encapsulation, we can guide the LLM to generate the corresponding CuTe APIs through simple few-shot examples, thereby extending support to a broader range of advanced GPU architectures.

\section{Evaluation}
\subsection{Experiment Setup}
To validate the performance of the attention operator implementation generated by LLMs through \name, we conduct comprehensive evaluation across different GPU platforms, distinct LLMs, and attention variants. 
%experiments are organized around four main factors that influence performance.
 Additional experiments are shown in Appendix~\ref{sec:appendix} due to space limitations.

\textbf{GPU platforms}.  \name is test on GPUs with different architectures, including NVIDIA Ampere (A100)  and Turing (RTX8000, T4) architecture.

\textbf{LLMs}.  The proposed framework is validated on four SOTA LLMs, including the closed-source models of GPT-4o~\cite{gpt4o} and Claude 3.5 sonnet~\cite{claude35}, as well as the open-source models of DeepSeek-V3~\cite{deepseekv3} and DeepSeek-R1~\cite{deepseekr1}.

\textbf{Attention variants}.  Experiments are conducted on attention mechanisms commonly used by current LLMs, including MHA (GPT-style models), GQA (Llama 3.1~\cite{llama31}, Qwen2.5~\cite{qwen25technicalreport}), MQA (Falcon~\cite{falcon}, StarCoder~\cite{li2023starcodersourceyou}), and MLA (DeepSeek-V2, DeepSeek-V3).

\textbf{Comparison Baselines}. Implementation generated by our approach is compared with FlashAttention official library (v2.7.3 on A100 and v1.0.9 on RTX8000, as FlashAttention v2 is not available on Turing architecture), NVIDIA official deep learning library cuDNN~\cite{cudnn}, FlexAttention official library~\cite{dong2024flex}, 

% We found that LLMs, when activated by vanilla prompts, are relatively adept at generating attention operators using PyTorch, but they struggle to produce higher-performance attention operators through CUDA code. Therefore, we directly utilized the torch operators generated by LLMs as the optimal code that vanilla prompts can produce.

\textbf{Benchmark setting}. We followed the settings in the FlashAttention\cite{dao2022flashattention, dao2024flashattention-2}, the sequence length varies from 512, 1k, ..., 16k, and the batch size is adjusted such that the total number of tokens remains 16k. 
We set the hidden dimension to 2048, and the head dimension to be either 64 or 128 (i.e., 32 heads or 16 heads).
The FLOPs are calculated as: $4\,*\,\mathrm{seqlen}^2\,*\, \mathrm{head\,\,dimension}\,*\,\mathrm{number\,\,of\,\,heads}$. For MLA, we utilize the dimensions specified in DeepSeek-V3, which consist of an embedding dimension of 128 and a RoPE dimension of 64.
We also test attention variants with and without causal masks.% we divide this number by 2 to account for the fact that approximately only half of the entries are calculated.
\begin{table*}[]
\centering
\resizebox{\textwidth}{!}{
\begin{tabular}{c|c|l|rrrrrr|rrrrrr}
\toprule
 \multicolumn{3}{c|}{Head Dimension}        & \multicolumn{6}{c|}{64}                                                                             & \multicolumn{6}{c}{128}                                                                             \\ \midrule
                                  \multicolumn{3}{c|}{Sequence Length}      & 512            & 1k             & 2k             & 4k             & 8k             & 16k            & 512            & 1k             & 2k             & 4k             & 8k             & 16k            \\ \midrule

\multirow{18}{*}{\begin{tabular}[c]{@{}c@{}}A100 \end{tabular}}           & \multirow{6}{*}{\begin{tabular}[c]{@{}c@{}c@{}c@{}}MHA\\ w/ \\casual\\mask  \end{tabular}}
                                 & cuDNN         & 95.3           & 124.4          & 143.7          & 152.4          & 162.8          & 172.5          & 106.1          & 135.4          & 153.3          & 165.5          & 177.8          & 186.3          \\
                                 &                      & FlexAttention & 84.4          & 107.4          & 123.7          & 134.7          & 145.8          & 153.3         & 80.5    & 105.3          & 124.7          & 137.4          & 150.7     & 160.3       \\ 
                                 &                      & flash-attn v2 & 101.2          & 127.3          & 146.5          & 158.5          & 172.4          & 180.8          & 115.3          & 143.6          & 163.8          & \textbf{176.9} & 183.3          & \textbf{195.1} \\ 
                                 &               & DeepSeek-V3      & 7.6            & 7.7            & 5.5            & 6.7            & 7.5            & 7.7            & 14.3           & 14.9           & 10.7           & 12.9           & 14.5           & 14.9           \\
                                 &                     & \cellcolor[rgb]{0.925,0.925,0.925}DeepSeek-V3 + Ours         & \cellcolor[rgb]{0.925,0.925,0.925}\textbf{107.4} & \cellcolor[rgb]{0.925,0.925,0.925}\textbf{134.6} & \cellcolor[rgb]{0.925,0.925,0.925}\textbf{154.7} & \cellcolor[rgb]{0.925,0.925,0.925}\textbf{163.4} & \cellcolor[rgb]{0.925,0.925,0.925}\textbf{177.6} & \cellcolor[rgb]{0.925,0.925,0.925}\textbf{184.3} & \cellcolor[rgb]{0.925,0.925,0.925}\textbf{132.2} & \cellcolor[rgb]{0.925,0.925,0.925}\textbf{155.6} & \cellcolor[rgb]{0.925,0.925,0.925}\textbf{168.7} & \cellcolor[rgb]{0.925,0.925,0.925}176.2 & \cellcolor[rgb]{0.925,0.925,0.925}\textbf{184.9} & \cellcolor[rgb]{0.925,0.925,0.925}194.7          \\

                                 &                      &    \cellcolor[rgb]{0.925,0.925,0.925}     & \cellcolor[rgb]{0.925,0.925,0.925}\textbf{$\uparrow$14.13×} & \cellcolor[rgb]{0.925,0.925,0.925}\textbf{$\uparrow$17.48×} & \cellcolor[rgb]{0.925,0.925,0.925}\textbf{$\uparrow$28.13×} & \cellcolor[rgb]{0.925,0.925,0.925}\textbf{$\uparrow$24.39×} & \cellcolor[rgb]{0.925,0.925,0.925}\textbf{$\uparrow$23.68×}

  & \textbf{$\uparrow$34.94×}  & \textbf{$\uparrow$9.24×} & \textbf{$\uparrow$10.44×} & \textbf{$\uparrow$15.77×} & {$\uparrow$13.66×} & \textbf{$\uparrow$12.75×} & {$\uparrow$13.07×}
 \\
                                 \cmidrule{2-15} 
                                 & \multirow{6}{*}{\begin{tabular}[c]{@{}c@{}c@{}c@{}}GQA\\ w/ \\casual\\mask  \end{tabular}} 
                                & cuDNN         & 95.5           & 125.6          & 142.4          & 152.3          & 164.8          & 172.1          & 107.4          & 136.6          & 154.2          & 165.3          & 177.6          & 186.9          \\
                                 &                      & FlexAttention & 85.6          & 108.3          & 124.1          & 133.7          & 146.4          & 155.2          & 81.8          & 106.6          & 125.4          & 138.4 & 153.3 & 161.1 \\ 
                                 &                      & flash-attn v2 & 102.2          & 128.6          & 147.7          & 159.9          & 172.1          & 182.4          & 116.6          & 144.8          & 164.1          & \textbf{176.5} & \textbf{186.7} & \textbf{197.2} \\ 
                                 &                      & DeepSeek-V3      & 5.4            & 5.6            & 4.4            & 5.1            & 5.5            & 6.8            & 10.5           & 10.8           & 8.5            & 9.9            & 10.9           & 11.4           \\
                                 &                      & \cellcolor[rgb]{0.925,0.925,0.925} DeepSeek-V3 + Ours         & \cellcolor[rgb]{0.925,0.925,0.925}\textbf{107.2} & \cellcolor[rgb]{0.925,0.925,0.925}\textbf{133.5} & \cellcolor[rgb]{0.925,0.925,0.925}\textbf{154.7} & \cellcolor[rgb]{0.925,0.925,0.925}\textbf{163.7} & \cellcolor[rgb]{0.925,0.925,0.925}\textbf{179.8} & \cellcolor[rgb]{0.925,0.925,0.925}\textbf{185.9} & \cellcolor[rgb]{0.925,0.925,0.925}\textbf{131.1} & \cellcolor[rgb]{0.925,0.925,0.925}\textbf{154.2} & \cellcolor[rgb]{0.925,0.925,0.925}\textbf{167.5} & \cellcolor[rgb]{0.925,0.925,0.925}175.5 & \cellcolor[rgb]{0.925,0.925,0.925}184.4 & \cellcolor[rgb]{0.925,0.925,0.925}195.6 \\

                                 &                      & \cellcolor[rgb]{0.925,0.925,0.925}      & \cellcolor[rgb]{0.925,0.925,0.925}\textbf{$\uparrow$19.85×} & \cellcolor[rgb]{0.925,0.925,0.925}\textbf{$\uparrow$23.84×} & \cellcolor[rgb]{0.925,0.925,0.925}\textbf{$\uparrow$35.16×} & \cellcolor[rgb]{0.925,0.925,0.925}\textbf{$\uparrow$32.10×} & \cellcolor[rgb]{0.925,0.925,0.925}\textbf{$\uparrow$32.69×} & \cellcolor[rgb]{0.925,0.925,0.925}\textbf{$\uparrow$27.34×} & \cellcolor[rgb]{0.925,0.925,0.925}\textbf{$\uparrow$12.49×} & \cellcolor[rgb]{0.925,0.925,0.925}\textbf{$\uparrow$14.28×} & \cellcolor[rgb]{0.925,0.925,0.925}\textbf{$\uparrow$19.71×} & \cellcolor[rgb]{0.925,0.925,0.925}{$\uparrow$17.73×} & \cellcolor[rgb]{0.925,0.925,0.925}{$\uparrow$16.92×} & \cellcolor[rgb]{0.925,0.925,0.925}{$\uparrow$17.16×} \\

                                 \cmidrule{2-15} 
                                 & \multirow{6}{*}{\begin{tabular}[c]{@{}c@{}c@{}c@{}}MQA\\ w/ \\casual\\mask  \end{tabular}} 
                                 & cuDNN         & 95.1           & 125.4          & 143.6          & 152.8          & 163.4          & 172.7          & 108.8          & 136.2          & 154.3          & 165.6          & 178.7          & 185.5          \\
                                 &                      & FlexAttention & 85.9          & 108.4          & 124.4          & 133.9          & 145.6          & 154.7          & 82.7          & 106.9          & 125.7          & 137.9 & 151.6 & 161.8 \\ 
                                 &                      & flash-attn v2 & 103.4          & 129.6          & 147.1          & 159.9          & 172.5          & 183.7          & 116.1          & 144.5          & 164.4          & \textbf{176.9} & \textbf{185.1} & \textbf{196.4} \\ 
                                 &               & DeepSeek-V3      & 7.7            & 8.1            & 5.8            & 6.9            & 7.5            & 7.7            & 14.7           & 15.1           & 10.9           & 13.2           & 14.7           & 15.1           \\
                                 &                      & \cellcolor[rgb]{0.925,0.925,0.925}   DeepSeek-V3 + Ours          & \cellcolor[rgb]{0.925,0.925,0.925}\textbf{107.6} & \cellcolor[rgb]{0.925,0.925,0.925}\textbf{134.8} & \cellcolor[rgb]{0.925,0.925,0.925}\textbf{155.4} & \cellcolor[rgb]{0.925,0.925,0.925}\textbf{163.7} & \cellcolor[rgb]{0.925,0.925,0.925}\textbf{179.2} & \cellcolor[rgb]{0.925,0.925,0.925}\textbf{186.9} & \cellcolor[rgb]{0.925,0.925,0.925}\textbf{131.1} & \cellcolor[rgb]{0.925,0.925,0.925}\textbf{153.6} & \cellcolor[rgb]{0.925,0.925,0.925}\textbf{167.7} & \cellcolor[rgb]{0.925,0.925,0.925}175.8 & \cellcolor[rgb]{0.925,0.925,0.925}183.9 & \cellcolor[rgb]{0.925,0.925,0.925}193.4 \\
                                 &                      & \cellcolor[rgb]{0.925,0.925,0.925}         & \cellcolor[rgb]{0.925,0.925,0.925}\textbf{$\uparrow$13.97×} & \cellcolor[rgb]{0.925,0.925,0.925}\textbf{$\uparrow$16.64×} & \cellcolor[rgb]{0.925,0.925,0.925}\textbf{$\uparrow$26.79×} & \cellcolor[rgb]{0.925,0.925,0.925}\textbf{$\uparrow$23.72×} & \cellcolor[rgb]{0.925,0.925,0.925}\textbf{$\uparrow$23.89×} & \cellcolor[rgb]{0.925,0.925,0.925}\textbf{$\uparrow$24.27×} & \cellcolor[rgb]{0.925,0.925,0.925}\textbf{$\uparrow$8.92×} & \cellcolor[rgb]{0.925,0.925,0.925}\textbf{$\uparrow$10.17×} & \cellcolor[rgb]{0.925,0.925,0.925}\textbf{$\uparrow$15.39×} & \cellcolor[rgb]{0.925,0.925,0.925}{$\uparrow$13.32×} & \cellcolor[rgb]{0.925,0.925,0.925}{$\uparrow$12.51×} & \cellcolor[rgb]{0.925,0.925,0.925}{$\uparrow$12.81×} \\

                                 \midrule
\multirow{18}{*}{\begin{tabular}[c]{@{}c@{}}RTX8000\end{tabular}}        & \multirow{6}{*}{\begin{tabular}[c]{@{}c@{}c@{}c@{}}MHA\\ w/ \\casual\\mask  \end{tabular}} 
                                 & cuDNN         & 21.4  & 25.7           & 28.7           & 31.2           & 32.7           & 33.5           & 20.9           & 25.0             & 28.5           & 31.1           & 32.1           & 32.3           \\
                                 &                      & FlexAttention  & \textbf{30.4}           & \textbf{34.5}           & \textbf{39.7}           & \textbf{43.9}           & 46.6           & 47.7           & 20.9           & 24.5           & 28.7           & 31.6           & 33.7           & 34.1           \\ 
                                 &                      & flash-attn v1 & 18.1           & 17.9           & 24.3           & 26.8           & 31.1           & 33.7           & 11.7           & 16.8           & 18.4           & 19.1           & 21.5           & 21.9           \\ 
                                 &                      & DeepSeek-V3      & 2.6            & 2.5            & 1.9            & 2.4            & 2.6            & OOM            & 5.1            & 5.2            & 3.9            & 4.7            & 5.1            & 4.8            \\
                                 &                      & \cellcolor[rgb]{0.925,0.925,0.925} DeepSeek-V3 + Ours         & \cellcolor[rgb]{0.925,0.925,0.925}21.6 & \cellcolor[rgb]{0.925,0.925,0.925}29.6 & \cellcolor[rgb]{0.925,0.925,0.925}37.9 & \cellcolor[rgb]{0.925,0.925,0.925}43.5 & \cellcolor[rgb]{0.925,0.925,0.925}\textbf{47.8} & \cellcolor[rgb]{0.925,0.925,0.925}\textbf{49.9} & \cellcolor[rgb]{0.925,0.925,0.925}\textbf{32.3} & \cellcolor[rgb]{0.925,0.925,0.925}\textbf{33.1} & \cellcolor[rgb]{0.925,0.925,0.925}\textbf{39.1} & \cellcolor[rgb]{0.925,0.925,0.925}\textbf{41.2} & \cellcolor[rgb]{0.925,0.925,0.925}\textbf{45.0} & \cellcolor[rgb]{0.925,0.925,0.925}\textbf{45.6} \\
                                 &                      & \cellcolor[rgb]{0.925,0.925,0.925}         & \cellcolor[rgb]{0.925,0.925,0.925}{$\uparrow$8.31×} & \cellcolor[rgb]{0.925,0.925,0.925}{$\uparrow$11.84×} & \cellcolor[rgb]{0.925,0.925,0.925}{$\uparrow$19.95×} & \cellcolor[rgb]{0.925,0.925,0.925}{$\uparrow$18.13×} & \cellcolor[rgb]{0.925,0.925,0.925}\textbf{$\uparrow$18.38×} & \cellcolor[rgb]{0.925,0.925,0.925}- & \cellcolor[rgb]{0.925,0.925,0.925}\textbf{$\uparrow$6.33×} & \cellcolor[rgb]{0.925,0.925,0.925}\textbf{$\uparrow$6.37×} & \cellcolor[rgb]{0.925,0.925,0.925}\textbf{$\uparrow$10.03×} & \cellcolor[rgb]{0.925,0.925,0.925}\textbf{$\uparrow$8.77×} & \cellcolor[rgb]{0.925,0.925,0.925}\textbf{$\uparrow$8.82×} & \cellcolor[rgb]{0.925,0.925,0.925}\textbf{$\uparrow$9.50×} \\

                                 \cmidrule{2-15} 
                                 & \multirow{6}{*}{\begin{tabular}[c]{@{}c@{}c@{}c@{}}GQA\\ w/ \\casual\\mask  \end{tabular}} 
                                 & cuDNN         & 21.5  & 26.5           & 28.9           & 31.2           & 32.5           & 32.9           & 20.4           & 23.3           & 28.1           & 30.4           & 32.3           & 32.3           \\
                                 &                      & FlexAttention & \textbf{30.6}           & \textbf{34.3}           & \textbf{39.6}           & \textbf{44.2}           & 46.9           & 47.8           & 20.9           & 24.3           & 28.7           & 31.8           & 33.6           & 34.2           \\ 
                                 &                      & flash-attn v1 & 18.1           & 18.1           & 24.4           & 28.4           & 31.4           & 33.7           & 11.9           & 16.8           & 18.8           & 19.1           & 21.0           & 21.9           \\ 
                                 &                      & DeepSeek-V3      & 2.1            & 2.2            & 1.7            & 1.99           & 2.2            & OOM            & 4.2            & 4.3            & 3.4            & 3.9            & 4.2            & 4.0              \\
                                 &                      & \cellcolor[rgb]{0.925,0.925,0.925} DeepSeek-V3 + Ours          & \cellcolor[rgb]{0.925,0.925,0.925}{21.6} & \cellcolor[rgb]{0.925,0.925,0.925}{28.6} & \cellcolor[rgb]{0.925,0.925,0.925}{39.5} & \cellcolor[rgb]{0.925,0.925,0.925}{42.4} & \cellcolor[rgb]{0.925,0.925,0.925}\textbf{47.9} & \cellcolor[rgb]{0.925,0.925,0.925}\textbf{50.4} & \cellcolor[rgb]{0.925,0.925,0.925}\textbf{25.9} & \cellcolor[rgb]{0.925,0.925,0.925}\textbf{31.5} & \cellcolor[rgb]{0.925,0.925,0.925}\textbf{36.4} & \cellcolor[rgb]{0.925,0.925,0.925}\textbf{40.7} & \cellcolor[rgb]{0.925,0.925,0.925}\textbf{44.6} & \cellcolor[rgb]{0.925,0.925,0.925}\textbf{45.5} \\
                                 &                      & \cellcolor[rgb]{0.925,0.925,0.925}        & \cellcolor[rgb]{0.925,0.925,0.925}{$\uparrow$10.28×} & \cellcolor[rgb]{0.925,0.925,0.925}{$\uparrow$13.00×} & \cellcolor[rgb]{0.925,0.925,0.925}{$\uparrow$23.24×} & \cellcolor[rgb]{0.925,0.925,0.925}{$\uparrow$21.31×} & \cellcolor[rgb]{0.925,0.925,0.925}\textbf{$\uparrow$21.77×} & \cellcolor[rgb]{0.925,0.925,0.925}- & \cellcolor[rgb]{0.925,0.925,0.925}\textbf{$\uparrow$6.17×} & \cellcolor[rgb]{0.925,0.925,0.925}\textbf{$\uparrow$7.33×} & \cellcolor[rgb]{0.925,0.925,0.925}\textbf{$\uparrow$10.71×} & \cellcolor[rgb]{0.925,0.925,0.925}\textbf{$\uparrow$10.44×} & \cellcolor[rgb]{0.925,0.925,0.925}\textbf{$\uparrow$10.62×} & \cellcolor[rgb]{0.925,0.925,0.925}\textbf{$\uparrow$11.38×} \\

                                 \cmidrule{2-15} 
                                 & \multirow{6}{*}{\begin{tabular}[c]{@{}c@{}c@{}c@{}}MQA\\ w/ \\casual\\mask  \end{tabular}} 
                                 & cuDNN         & 21.8  & 26.3           & 30.2           & 31.7           & 32.7           & 33.1           & 20.5           & 25.4           & 29.3           & 30.9           & 31.6           & 31.9           \\
                                 &                      & FlexAttention & \textbf{30.7}           & \textbf{34.5}           & \textbf{39.8}           & \textbf{44.5}           & 46.8           & 47.5           & 20.2           & 23.9           & 28.5           & 32.4           & 33.8           & 34.4           \\
                                 &                      & flash-attn v1 & 18.1           & 18.3           & 25.4           & 30.1           & 31.2           & 34.2           & 11.9           & 16.8           & 18.8           & 19.1           & 21.2           & 22.0           \\
                                 &                      & DeepSeek-V3       & 2.6            & 2.6            & 1.9            & 2.5           & 2.7            & OOM            & 5.1            & 5.3            & 4.1            & 4.8            & 5.3            & 4.9              \\
                                 &                      & \cellcolor[rgb]{0.925,0.925,0.925} DeepSeek-V3 + Ours          & \cellcolor[rgb]{0.925,0.925,0.925}{22.1} & \cellcolor[rgb]{0.925,0.925,0.925}{29.8} & \cellcolor[rgb]{0.925,0.925,0.925}{39.1} & \cellcolor[rgb]{0.925,0.925,0.925}{43.3} & \cellcolor[rgb]{0.925,0.925,0.925}\textbf{47.8} & \cellcolor[rgb]{0.925,0.925,0.925}\textbf{50.6} & \cellcolor[rgb]{0.925,0.925,0.925}\textbf{26.1} & \cellcolor[rgb]{0.925,0.925,0.925}\textbf{32.5} & \cellcolor[rgb]{0.925,0.925,0.925}\textbf{36.4} & \cellcolor[rgb]{0.925,0.925,0.925}\textbf{40.7} & \cellcolor[rgb]{0.925,0.925,0.925}\textbf{43.7} & \cellcolor[rgb]{0.925,0.925,0.925}\textbf{45.1} \\
                                 &                      & \cellcolor[rgb]{0.925,0.925,0.925}       & \cellcolor[rgb]{0.925,0.925,0.925}{$\uparrow$8.50×} & \cellcolor[rgb]{0.925,0.925,0.925}{$\uparrow$11.46×} & \cellcolor[rgb]{0.925,0.925,0.925}{$\uparrow$20.58×} & \cellcolor[rgb]{0.925,0.925,0.925}{$\uparrow$17.32×} & \cellcolor[rgb]{0.925,0.925,0.925}\textbf{$\uparrow$17.70×} & \cellcolor[rgb]{0.925,0.925,0.925}- & \cellcolor[rgb]{0.925,0.925,0.925}\textbf{$\uparrow$5.12×} & \cellcolor[rgb]{0.925,0.925,0.925}\textbf{$\uparrow$6.13×} & \cellcolor[rgb]{0.925,0.925,0.925}\textbf{$\uparrow$8.88×} & \cellcolor[rgb]{0.925,0.925,0.925}\textbf{$\uparrow$8.48×} & \cellcolor[rgb]{0.925,0.925,0.925}\textbf{$\uparrow$8.25×} & \cellcolor[rgb]{0.925,0.925,0.925}\textbf{$\uparrow$9.2×} \\

                                  \midrule
                                 \midrule
\multirow{18}{*}{\begin{tabular}[c]{@{}c@{}}A100\end{tabular}}           & \multirow{6}{*}{\begin{tabular}[c]{@{}c@{}c@{}c@{}}MHA\\ w/o \\casual\\mask  \end{tabular}} & cuDNN         & 153.0            & 158.8          & 172.4          & 175.5          & 184.7          & 186.2          & 172.4          & 184.2          & 190.0            & 196.5          & 206.9          & 212.0            \\
        &                      & FlexAttention & 145.8          & 155.9          & 162.5          & 168.4          & 177.2          & 179.9          & 119.7 & 134.8 & 143.2 & 152.4 & 158.6 & 163.7 \\ 

        &                      & flash-attn v2 & 147.5          & 161.6          & 171.1          & 176.8          & 185.8          & 190.6          & \textbf{208.1} & \textbf{208.1} & \textbf{208.2} & \textbf{210.5} & \textbf{221.4} & \textbf{223.6} \\ 
        &                 & DeepSeek-V3  & 28.9           & 29.6           & 28.2           & 28.5           & 28.5           & 29.6           & 51.6           & 54.5           & 52.4           & 52.6           & 54.7           & 56.6           \\
       &                      & \cellcolor[rgb]{0.925,0.925,0.925} DeepSeek-V3 + Ours          & \cellcolor[rgb]{0.925,0.925,0.925}\textbf{164.0} & \cellcolor[rgb]{0.925,0.925,0.925}\textbf{175.6} & \cellcolor[rgb]{0.925,0.925,0.925}\textbf{181.8} & \cellcolor[rgb]{0.925,0.925,0.925}\textbf{191.0} & \cellcolor[rgb]{0.925,0.925,0.925}\textbf{200.6} & \cellcolor[rgb]{0.925,0.925,0.925}\textbf{201.8} & \cellcolor[rgb]{0.925,0.925,0.925}176.2 & \cellcolor[rgb]{0.925,0.925,0.925}194.1 & \cellcolor[rgb]{0.925,0.925,0.925}201.1 & \cellcolor[rgb]{0.925,0.925,0.925}205.6 & \cellcolor[rgb]{0.925,0.925,0.925}206.6 & \cellcolor[rgb]{0.925,0.925,0.925}207.2 \\
       &                      & \cellcolor[rgb]{0.925,0.925,0.925}       & \cellcolor[rgb]{0.925,0.925,0.925}\textbf{$\uparrow$5.67×} & \cellcolor[rgb]{0.925,0.925,0.925}\textbf{$\uparrow$5.93×} & \cellcolor[rgb]{0.925,0.925,0.925}\textbf{$\uparrow$6.45×} & \cellcolor[rgb]{0.925,0.925,0.925}\textbf{$\uparrow$6.70×} & \cellcolor[rgb]{0.925,0.925,0.925}\textbf{$\uparrow$7.03×} & \cellcolor[rgb]{0.925,0.925,0.925}\textbf{$\uparrow$6.82×} & \cellcolor[rgb]{0.925,0.925,0.925}{$\uparrow$3.41×} & \cellcolor[rgb]{0.925,0.925,0.925}{$\uparrow$3.56×} & \cellcolor[rgb]{0.925,0.925,0.925}{$\uparrow$3.84×} & \cellcolor[rgb]{0.925,0.925,0.925}{$\uparrow$3.91×} & \cellcolor[rgb]{0.925,0.925,0.925}{$\uparrow$3.81×} & \cellcolor[rgb]{0.925,0.925,0.925}{$\uparrow$3.73×} \\

        \cmidrule{2-15} 
        & \multirow{6}{*}{\begin{tabular}[c]{@{}c@{}c@{}c@{}}GQA\\ w/o \\casual\\mask  \end{tabular}} 
        & cuDNN         & 148.8          & 159.1          & 164.3          & 172.3          & 180.5          & 186.7          & 172.3          & 183.5          & 189.6          & 193.6          & 206.9          & 211.1          \\
        &                      & FlexAttention & 146.2          & 156.3          & 161.6          & 172.8          & 176.9          & 180.5          & 121.1 & 135.1 & 143.5 & 155.2 & 160.5 & 165.5 \\ 
        
        &                      & flash-attn v2 & 148.4          & 161.2          & 168.5          & 175.7          & 186.8          & 190.1          & \textbf{176.9} & \textbf{192.5} & \textbf{200.0} & \textbf{210.4} & \textbf{218.4} & \textbf{223.8} \\ 
        &                 & DeepSeek-V3  & 14.3           & 14.8           & 11.7           & 13.4           & 14.9           & 19.8           & 27.5           & 28.7           & 23.1           & 26.3           & 29.1           & 30.8           \\
        &                      & \cellcolor[rgb]{0.925,0.925,0.925} DeepSeek-V3 + Ours            & \cellcolor[rgb]{0.925,0.925,0.925}\textbf{156.9} & \cellcolor[rgb]{0.925,0.925,0.925}\textbf{168.4} & \cellcolor[rgb]{0.925,0.925,0.925}\textbf{177.6} & \cellcolor[rgb]{0.925,0.925,0.925}\textbf{186.9} & \cellcolor[rgb]{0.925,0.925,0.925}\textbf{195.1} & \cellcolor[rgb]{0.925,0.925,0.925}\textbf{202.7} & \cellcolor[rgb]{0.925,0.925,0.925}169.6 & \cellcolor[rgb]{0.925,0.925,0.925}180.4 & \cellcolor[rgb]{0.925,0.925,0.925}186.2 & \cellcolor[rgb]{0.925,0.925,0.925}196.7 & \cellcolor[rgb]{0.925,0.925,0.925}204.8 & \cellcolor[rgb]{0.925,0.925,0.925}206.9 \\
        &                      & \cellcolor[rgb]{0.925,0.925,0.925}      & \cellcolor[rgb]{0.925,0.925,0.925}\textbf{$\uparrow$10.97×} & \cellcolor[rgb]{0.925,0.925,0.925}\textbf{$\uparrow$11.38×} & \cellcolor[rgb]{0.925,0.925,0.925}\textbf{$\uparrow$15.18×} & \cellcolor[rgb]{0.925,0.925,0.925}\textbf{$\uparrow$13.95×} & \cellcolor[rgb]{0.925,0.925,0.925}\textbf{$\uparrow$13.09×} & \cellcolor[rgb]{0.925,0.925,0.925}\textbf{$\uparrow$10.24×} & \cellcolor[rgb]{0.925,0.925,0.925}{$\uparrow$6.17×} & \cellcolor[rgb]{0.925,0.925,0.925}{$\uparrow$6.29×} & \cellcolor[rgb]{0.925,0.925,0.925}{$\uparrow$8.06×} & \cellcolor[rgb]{0.925,0.925,0.925}{$\uparrow$7.48×} & \cellcolor[rgb]{0.925,0.925,0.925}{$\uparrow$7.04×} & \cellcolor[rgb]{0.925,0.925,0.925}{$\uparrow$6.72×} \\

        \cmidrule{2-15} 
        & \multirow{6}{*}{\begin{tabular}[c]{@{}c@{}c@{}c@{}}MQA\\ w/o \\casual\\mask  \end{tabular}} 
                                  & cuDNN         & 149.1          & 159.6          & 164.7          & 173.5          & 180.4          & 185.5          & 172.5          & 184.8          & 189.9          & 197.5          & 208.1          & 211.4          \\
                                 &                      & FlexAttention & 145.8          & 155.8          & 161.5          & 173.4          & 176.3          & 180.4          & 119.5 & 135.6 & 143.5 & 152.5 & 158.4 & 164.9 \\ 
                                 &                      & flash-attn v2 & 149.6          & 162.1          & 168.2          & 175.8          & 185.9          & 190.1          & \textbf{178.2} & \textbf{193.2} & \textbf{200.7} & \textbf{208.8} & \textbf{219.1} & \textbf{225.8} \\ 
        &                         & DeepSeek-V3       & 29.2           & 30.7           & 19.9           & 25.4           & 29.1           & 30.2           & 54.1           & 56.2           & 38.4           & 48.9           & 55.6           & 58.9           \\
        &                      & \cellcolor[rgb]{0.925,0.925,0.925} DeepSeek-V3 + Ours          & \cellcolor[rgb]{0.925,0.925,0.925}\textbf{156.5} & \cellcolor[rgb]{0.925,0.925,0.925}\textbf{168.4} & \cellcolor[rgb]{0.925,0.925,0.925}\textbf{176.8} & \cellcolor[rgb]{0.925,0.925,0.925}\textbf{186.7} & \cellcolor[rgb]{0.925,0.925,0.925}\textbf{196.7} & \cellcolor[rgb]{0.925,0.925,0.925}\textbf{201.2} & \cellcolor[rgb]{0.925,0.925,0.925}169.2 & \cellcolor[rgb]{0.925,0.925,0.925}180.5 & \cellcolor[rgb]{0.925,0.925,0.925}187.6 & \cellcolor[rgb]{0.925,0.925,0.925}197.9 & \cellcolor[rgb]{0.925,0.925,0.925}204.6 & \cellcolor[rgb]{0.925,0.925,0.925}207.8 \\
        &                      & \cellcolor[rgb]{0.925,0.925,0.925}     & \cellcolor[rgb]{0.925,0.925,0.925}\textbf{$\uparrow$5.36×} & \cellcolor[rgb]{0.925,0.925,0.925}\textbf{$\uparrow$5.49×} & \cellcolor[rgb]{0.925,0.925,0.925}\textbf{$\uparrow$8.88×} & \cellcolor[rgb]{0.925,0.925,0.925}\textbf{$\uparrow$7.35×} & \cellcolor[rgb]{0.925,0.925,0.925}\textbf{$\uparrow$6.76×} & \cellcolor[rgb]{0.925,0.925,0.925}\textbf{$\uparrow$6.66×} & \cellcolor[rgb]{0.925,0.925,0.925}{$\uparrow$3.13×} & \cellcolor[rgb]{0.925,0.925,0.925}{$\uparrow$3.21×} & \cellcolor[rgb]{0.925,0.925,0.925}{$\uparrow$4.89×} & \cellcolor[rgb]{0.925,0.925,0.925}{$\uparrow$4.05×} & \cellcolor[rgb]{0.925,0.925,0.925}{$\uparrow$3.68×} & \cellcolor[rgb]{0.925,0.925,0.925}{$\uparrow$3.53×} \\

        \midrule
\multirow{18}{*}{\begin{tabular}[c]{@{}c@{}}RTX8000\end{tabular}}        & \multirow{6}{*}{\begin{tabular}[c]{@{}c@{}c@{}c@{}}MHA\\ w/o \\casual\\mask  \end{tabular}}
                                 & cuDNN         & 29.2           & 32.2           & 33.4           & 33.6           & 33.7           & 33.4           & 26.5           & 30.9           & 32.2           & 32.6           & 32.8           & 32.6           \\
                                 &                      & FlexAttention & \textbf{40.5}           & \textbf{44.7}           & \textbf{47.5}           & \textbf{49.4}           & \textbf{50.1}           & \textbf{50.4}           & 29.4           & 32.5           & 33.2           & 34.7           & 35.3           & 35.5           \\
                                 
                                 &                      & flash-attn v2 & 28.5           & 25.9           & 34.3           & 35.1           & 35.4           & 36.2           & 17.2           & 20.4           & 21.2           & 21.0           & 22.0           & 22.4           \\
                                 &                       & DeepSeek-V3      & 8.3            & 8.5            & 6.6            & 8.4            & 9.3            & 8.9            & 15.7           & 16.6           & 13.4           & 16.5           & 17.9           & 15.7           \\
                                 &                      & \cellcolor[rgb]{0.925,0.925,0.925} DeepSeek-V3 + Ours          & \cellcolor[rgb]{0.925,0.925,0.925}{34.4} & \cellcolor[rgb]{0.925,0.925,0.925}{40.5} & \cellcolor[rgb]{0.925,0.925,0.925}{44.4} & \cellcolor[rgb]{0.925,0.925,0.925}{45.9} & \cellcolor[rgb]{0.925,0.925,0.925}{45.5} & \cellcolor[rgb]{0.925,0.925,0.925}{45.1} & \cellcolor[rgb]{0.925,0.925,0.925}\textbf{38.5} & \cellcolor[rgb]{0.925,0.925,0.925}\textbf{42.9} & \cellcolor[rgb]{0.925,0.925,0.925}\textbf{44.9} & \cellcolor[rgb]{0.925,0.925,0.925}\textbf{45.8} & \cellcolor[rgb]{0.925,0.925,0.925}\textbf{44.4} & \cellcolor[rgb]{0.925,0.925,0.925}\textbf{43.9} \\
                                 &                      & \cellcolor[rgb]{0.925,0.925,0.925}        & \cellcolor[rgb]{0.925,0.925,0.925}{$\uparrow$4.14×} & \cellcolor[rgb]{0.925,0.925,0.925}{$\uparrow$4.76×} & \cellcolor[rgb]{0.925,0.925,0.925}{$\uparrow$6.73×} & \cellcolor[rgb]{0.925,0.925,0.925}{$\uparrow$5.46×} & \cellcolor[rgb]{0.925,0.925,0.925}{$\uparrow$4.89×} & \cellcolor[rgb]{0.925,0.925,0.925}{$\uparrow$5.07×} & \cellcolor[rgb]{0.925,0.925,0.925}\textbf{$\uparrow$2.45×} & \cellcolor[rgb]{0.925,0.925,0.925}\textbf{$\uparrow$2.58×} & \cellcolor[rgb]{0.925,0.925,0.925}\textbf{$\uparrow$3.35×} & \cellcolor[rgb]{0.925,0.925,0.925}\textbf{$\uparrow$2.78×} & \cellcolor[rgb]{0.925,0.925,0.925}\textbf{$\uparrow$2.48×} & \cellcolor[rgb]{0.925,0.925,0.925}\textbf{$\uparrow$2.80×} \\
                                 \cmidrule{2-15} 
                                 & \multirow{6}{*}{\begin{tabular}[c]{@{}c@{}c@{}c@{}}GQA\\ w/o \\casual\\mask  \end{tabular}} 
                                 & cuDNN         & 26.8           & 31.2           & 32.1           & 32.9           & 33.7           & 33.8           & 26.3           & 29.7           & 32.1           & 32.6           & 33.1           & 32.8           \\
                                 &                      & FlexAttention & \textbf{39.2}           & \textbf{43.9}           & \textbf{47.3}           & \textbf{49.4}           & \textbf{49.8}           & \textbf{50.2}           & 29.5           & 32.2           & 33.4           & 34.9           & 35.2           & 35.1           \\ 
                                 
                                 &                      & flash-attn v1 & 28.9           & 26.5           & 33.7           & 35.7           & 35.8           & 37.1           & 17.4           & 20.5           & 21.1           & 21.1           & 22.1           & 22.5           \\ 
                                 &                      & DeepSeek-V3      & 5.1            & 5.1            & 4.4            & 5.1            & 5.5            & OOM            & 10.1           & 10.3           & 8.8            & 10.1           & 10.8           & 9.9            \\
                                 &                      & \cellcolor[rgb]{0.925,0.925,0.925} DeepSeek-V3 + Ours          & \cellcolor[rgb]{0.925,0.925,0.925}{30.3} & \cellcolor[rgb]{0.925,0.925,0.925}{39.3} & \cellcolor[rgb]{0.925,0.925,0.925}{43.9} & \cellcolor[rgb]{0.925,0.925,0.925}{45.4} & \cellcolor[rgb]{0.925,0.925,0.925}{45.8} & \cellcolor[rgb]{0.925,0.925,0.925}{46.1} & \cellcolor[rgb]{0.925,0.925,0.925}\textbf{37.4} & \cellcolor[rgb]{0.925,0.925,0.925}\textbf{41.9} & \cellcolor[rgb]{0.925,0.925,0.925}\textbf{43.3} & \cellcolor[rgb]{0.925,0.925,0.925}\textbf{45.1} & \cellcolor[rgb]{0.925,0.925,0.925}\textbf{45.1} & \cellcolor[rgb]{0.925,0.925,0.925}\textbf{45.2} \\
                                 &                      & \cellcolor[rgb]{0.925,0.925,0.925}        & \cellcolor[rgb]{0.925,0.925,0.925}{$\uparrow$5.94×} & \cellcolor[rgb]{0.925,0.925,0.925}{$\uparrow$7.71×} & \cellcolor[rgb]{0.925,0.925,0.925}{$\uparrow$9.98×} & \cellcolor[rgb]{0.925,0.925,0.925}{$\uparrow$8.90×} & \cellcolor[rgb]{0.925,0.925,0.925}{$\uparrow$8.33×} & \cellcolor[rgb]{0.925,0.925,0.925}- & \cellcolor[rgb]{0.925,0.925,0.925}\textbf{$\uparrow$3.70×} & \cellcolor[rgb]{0.925,0.925,0.925}\textbf{$\uparrow$4.07×} & \cellcolor[rgb]{0.925,0.925,0.925}\textbf{$\uparrow$4.92×} & \cellcolor[rgb]{0.925,0.925,0.925}\textbf{$\uparrow$4.47×} & \cellcolor[rgb]{0.925,0.925,0.925}\textbf{$\uparrow$4.18×} & \cellcolor[rgb]{0.925,0.925,0.925}\textbf{$\uparrow$4.57×} \\

                                 \cmidrule{2-15} 
                                 & \multirow{6}{*}{\begin{tabular}[c]{@{}c@{}c@{}c@{}}MQA\\ w/o \\casual\\mask  \end{tabular}} 
                                 
                                & cuDNN         & 27.2           & 31.1           & 32.6           & 33.4           & 33.8           & 34.1           & 25.1           & 29.3           & 31.2           & 32.4           & 32.9           & 32.5           \\
                                 &                      & FlexAttention  & \textbf{39.4}           & \textbf{44.1}           & \textbf{46.7}           & \textbf{49.3}           & \textbf{49.6}           & \textbf{50.2}           & 29.4           & 32.7           & 33.5           & 34.6           & 34.9           & 35.3           \\ 
                                 
                                 &                      & flash-attn v1 & 28.8           & 26.2           & 34.5           & 35.7           & 35.7           & 37.2           & 17.4           & 20.5           & 21.3           & 21.1           & 22.1           & 22.4           \\ 
                                 &                      & DeepSeek-V3      & 8.6            & 8.6            & 6.6            & 8.4            & 9.3            & OOM            & 16.7           & 17.2           & 13.2           & 16.2           & 17.8           & 15.8           \\
                                 &                      & \cellcolor[rgb]{0.925,0.925,0.925} DeepSeek-V3 + Ours         & \cellcolor[rgb]{0.925,0.925,0.925}{29.8} & \cellcolor[rgb]{0.925,0.925,0.925}{39.1} & \cellcolor[rgb]{0.925,0.925,0.925}{44.4} & \cellcolor[rgb]{0.925,0.925,0.925}{45.3} & \cellcolor[rgb]{0.925,0.925,0.925}{45.8} & \cellcolor[rgb]{0.925,0.925,0.925}{45.9} & \cellcolor[rgb]{0.925,0.925,0.925}\textbf{36.5} & \cellcolor[rgb]{0.925,0.925,0.925}\textbf{41.6} & \cellcolor[rgb]{0.925,0.925,0.925}\textbf{43.4} & \cellcolor[rgb]{0.925,0.925,0.925}\textbf{44.9} & \cellcolor[rgb]{0.925,0.925,0.925}\textbf{44.8} & \cellcolor[rgb]{0.925,0.925,0.925}\textbf{45.1} \\
                                 &                      & \cellcolor[rgb]{0.925,0.925,0.925}       & \cellcolor[rgb]{0.925,0.925,0.925}{$\uparrow$3.47×} & \cellcolor[rgb]{0.925,0.925,0.925}{$\uparrow$4.55×} & \cellcolor[rgb]{0.925,0.925,0.925}{$\uparrow$6.73×} & \cellcolor[rgb]{0.925,0.925,0.925}{$\uparrow$5.39×} & \cellcolor[rgb]{0.925,0.925,0.925}{$\uparrow$4.92×} & \cellcolor[rgb]{0.925,0.925,0.925}- & \cellcolor[rgb]{0.925,0.925,0.925}\textbf{$\uparrow$2.19×} & \cellcolor[rgb]{0.925,0.925,0.925}\textbf{$\uparrow$2.42×} & \cellcolor[rgb]{0.925,0.925,0.925}\textbf{$\uparrow$3.29×} & \cellcolor[rgb]{0.925,0.925,0.925}\textbf{$\uparrow$2.77×} & \cellcolor[rgb]{0.925,0.925,0.925}\textbf{$\uparrow$2.52×} & \cellcolor[rgb]{0.925,0.925,0.925}\textbf{$\uparrow$2.85×} \\

                                 \bottomrule
\end{tabular}
}
\caption{Performance (TFLOPS) comparison across different sequence length, attention operators, GPUs, and the presence or absence of causal masks. OOM means out of memory. }%DeepSeek-V3 is used as the baseline model and we evaluate the speedup compared to the LLM's own capabilities. Additionally, we compare several representative handcraft libraries.}
\label{table:performance}
\end{table*}

\subsection{Overall Performance}
Table \ref{table:performance} demonstrates \name's performance across GPU architectures and common attention operators, while Table \ref{tbl:mla} specifically evaluates the novel MLA operator.
The results demonstrate \name's capability to empower DeepSeek-V3 in generating high-performance implementations for diverse attention operators across GPU platforms.

% \textbf{Compared to vanilla LLM.} \name outperforms the results generated by the vanilla LLM across both GPUs, all operators, all dimensions, and whether with causal mask, achieving a maximum improvement of $34.94\times$. It is noteworthy that the attention implementation generated by the vanilla prompt results in out-of-memory (OOM in table) errors when the sequence length is large, whereas ours do not encounter this issue. This advantage is partly attributed to the benefits of operator fusion and also serves as evidence that our method is capable of generating more advanced operator optimization techniques. 

\textbf{Consistent superiority compared with vanilla LLM}. \name achieves consistent superiority over vanilla LLM implementations across all tested scenarios (GPUs, operators, dimensions, and causal masking configurations), with peak speedups reaching $35.16\times$ in common attention operators and $9.41\times$ in MLA.
Notably, vanilla LLM-generated implementations even trigger out-of-memory (OOM) errors at large sequence lengths.
The results show our method can implement fused attention operators with less memory resource requirement and higher computation efficiency.

% \textbf{Compared to SOTA handcraft libraries}. More importantly, across most dimensions and operators, our approach achieves performance comparable to the handcraft libraries, with peak performance reaching up to $218\% (45.82/21.02)$. Moreover, the development cost of our method is significantly lower than that of handcrafted libraries. And it is similar for MLA operator, which achieves up to $8.16\times$ performance compared to vanilla LLM and $2.31\times(198.4/85.7)$ compared to cuDNN. We also conducted tests on the attention dimensions within existing state-of-the-art (SOTA) LLMs. The results demonstrate that our method retains the same advantages under real-world LLM dimensions. Further details and comprehensive analysis can be found in Appendix \ref{appendix:llm}.

\textbf{Competitive or superior performance compared with SOTA handcrafted libraries}. 
\name, which requires substantially lower development effort, demonstrates superior performance on the novel MLA operator with $2.15\times$ speedup (175.9 vs 81.7 TFLOPS) against SOTA handcrafted library, cuDNN, while achieving comparable or better performance even on established attention operators (like MHA, GQA and MQA) well-supported by existing optimized libraries.
Besides, our method maintains performance advantages over handcrafted libraries under real-world LLM-scale attention workloads (Appendix \ref{appendix:llm}), demonstrating strong generalizability and practical applicability in production environments.

\textbf{Generality across multiple generations of GPUs}.
\name demonstrates broad generality, being applicable to a wide range of GPUs, from the newer Ampere architecture to the older Turing architecture, the latter of which is not supported by official FlashAttention v2. This highlights \name's generality to reduce development overhead for operators across different GPU generations.
%\textbf{Compatibility support for multiple generations of GPUs}.
%\name can be compatible with a wide range of GPUs, spanning both the newer Ampere architecture and the older Turing architecture, while the latter is currently not supported by the official FlashAttention 2. This highlights the capability of \name to reduce the development overhead of operators to a certain extent. 
%Recently, DeepSeek V3 has significantly reduced the training and inference costs of LLMs, greatly enhancing the practical value of cost-efficient GPUs such as the RTX 8000. \name, with its inherent adaptability to diverse GPU architectures, naturally aligns with this trend, thereby demonstrating the practical significance of our approach.

\textbf{Outstanding performance in long-context scenarios with causal mask}.
\name performs exceptionally well in long-context scenarios, particularly when using a causal mask that is commonly employed in modern decoder-only LLMs ($34.94\times$ speedup on 16k tokens on A100).
%For instance, on an A100 GPU, we achieve a remarkable $34.94\times$ performance speedup on sequences as long as 16k tokens.
%Furthermore, on the RTX 8000, vanilla LLMs often encounter out-of-memory errors when handling long sequences, highlighting the superiority of our method in managing such lengths.
% As long-context capabilities become increasingly crucial in the latest LLMs, our approach's ability to efficiently process these sequences with a causal mask positions it as a significant advantage in contemporary model development.
Our method's efficient long-context sequences processing with causal mask offers a critical advantage in modern LLMs development.
%This advantage stems partly from the complexity benefits of the fused attention operator, but more significantly, it is because our method can activate the LLM to generate such a fusion on high-performance computing units. Conventional methods struggle to achieve this and often rely on existing libraries to implement suboptimal versions of the attention operator.

\begin{table}[htbp]
\resizebox{\linewidth}{!}{
\begin{tabular}{l|rrrrrr}
\toprule
Sequence Length    & 512   & 1k    & 2k    & 4k    & 8k    & 16k   \\ \midrule

torch & 22.9  & 28.7  & 21.7  & 26.7  & 32.9  & 35.1  \\ 
cuDNN       & 35.5  & 48.6  & 61.1  & 70.3  & 77.3  & 81.7 \\ 
DeepSeek-V3    & 17.7 & 18.5 & 13.5 & 16.1 & 18.2 & 18.7  \\ 
\rowcolor[rgb]{0.925,0.925,0.925} DeepSeek-V3 + Ours        & \textbf{50.6} & \textbf{78.6} & \textbf{108.2} & \textbf{138.6} & \textbf{164.3} & \textbf{175.9}
 \\
            \rowcolor[rgb]{0.925,0.925,0.925} & \textbf{$\uparrow$2.86×} & \textbf{$\uparrow$4.25×} & \textbf{$\uparrow$8.01×} & \textbf{$\uparrow$8.61×} & \textbf{$\uparrow$9.03×} & \textbf{$\uparrow$9.41×}
  \\ 
  \cmidrule{1-7}
\end{tabular}
}
\caption{Performance (TFLOPS) comparison of MLA with causal mask and head dimension 128 on A100 GPU. The MLA configuration parameters and ``torch'' implementation are extracted from the DeepSeek-V3 open-source code.}
% \caption{MLA with causal mask performance (TFLOPS) compared with handcraft libraries on A100 GPU under the configuration of head dimension 128, RoPE dimension 64 with causal mask.}
\label{tbl:mla}
\end{table}

%If a paper is accepted, we strongly encourage the publication of software and data with the
%camera-ready version of the paper whenever appropriate. %This can be
%done by including a URL in the camera-ready copy. However, \textbf{do not}
%include URLs that reveal your institution or identity in your
%submission for review. Instead, provide an anonymous URL or upload
%the material as ``Supplementary Material'' into the OpenReview reviewing
%system. Note that reviewers are not required to look at this material
%when writing their review.

% \section*{Acknowledgements}

% \textbf{Do not} include acknowledgements in the initial version of
% the paper submitted for blind review.

% If a paper is accepted, the final camera-ready version can (and
% usually should) include acknowledgements.  Such acknowledgements
% should be placed at the end of the section, in an unnumbered section
% that does not count towards the paper page limit. Typically, this will 
% include thanks to reviewers who gave useful comments, to colleagues 
% who contributed to the ideas, and to funding agencies and corporate 
% sponsors that provided financial support.
% \usepackage{multirow}

\subsection{Ablation}
\textbf{Impact of Different LLMs}. To demonstrate the robustness of \name, we conduct ablation experiments on various mainstream LLMs, including GPT-4o, Claude 3.5, DeepSeek-V3 and DeepSeek-R1, as shown in Table \ref{tbl:mha}.
After applying \name, nearly all LLMs can generate high-performance attention operators, which shows the strong versatility of our method.
%demonstrated significant improvements in code generation capabilities.
Notably, DeepSeek-R1 achieves the highest performance,
% metrics, surpassing all other LLMs in terms of code generation quality. 
while GPT-4o struggles to translate correct CuTe code, potentially due to limitations in its training corpus (GPT-4o is the earliest one within these LLMs).
However, it can still generate the TL Code, while the backend translation is handled by DeepSeek-V3. %This approach successfully produces high-performance attention operator code, demonstrating the strong versatility of our method even for LLMs with less comprehensive corpus coverage.
\begin{table}[]
\centering

\resizebox{\linewidth}{!}{
\begin{tabular}{l|r|r|r}
\toprule
LLM-TL            & Seq = 4k & Seq = 8k &Seq = 16k\\ \midrule
w/ GPT-4o      &    -     &   -    & -\\ 
w/ GPT-4o + DeepSeek-V3&  165.5 &   171.9 & 178.5   \\ 
w/ Claude 3.5  &  175.2   & 179.4   &    181.3   \\
w/ DeepSeek-V3 &  175.5   &  179.3  &   185.5    \\
w/ DeepSeek-R1 &  176.2  &  184.9   &   194.7    \\ \bottomrule
\end{tabular}
}
\caption{Performance (TFLOPS) comparison of MHA with causal mask and head dimension 128 generated by our methods with different LLMs on A100 GPU.}
\label{tbl:mha}
\end{table}

\begin{table}[]
\centering{
\resizebox{0.7\linewidth}{!}{
\begin{tabular}{l|lr}
\toprule
\multirow{1}{*}{}
                  & Time     & TFLOPS \\ \midrule
%Junior Developer  & 20 days      &  2.04   \\
Human Expert  & $\sim$ months    &  162.7        \\
\name        & 10 mins   & 175.6            \\ \bottomrule
\end{tabular}}
}
\caption{Comparison of MHA development cost with human expert on A100 GPU under the configuration of head dimension 64, sequence length 1024.}
\label{tab:develop_ablation}
\vspace{-10pt}
\end{table}

\textbf{Impact of Different Prompts}. 
% To validate the effectiveness of \name, we compare our method with Claude 3.5 and DeepSeek-V3 utilize CoT~\cite{wei2022cot}, as shown in Table \ref{tbl:cot}.
% the performance of attention implementation based on the low-level CUDA language 
% presents a comparative performance analysis of FlashAttention operator implementations between \name and CoT. Different with Table \ref{table:performance}, here the LLMs all generate CUDA Code rather than torch.
To demonstrate the efficacy of \name, we compare its performance with other LLMs using CoT for CUDA-based attention operator generation (notably, vanilla DeepSeek-V3 in Table \ref{table:performance} utilizes Torch).
Results in Table \ref{tbl:cot} demonstrate that while LLMs with CoT can only generate basic CUDA implementation, \name produces high-performance implementations based on CuTe. Performance evaluation reveals that \name achieves up to $895\times$ speedup compared to CoT.

\textbf{Impact on Development Cost}. 
Table \ref{tab:develop_ablation} compares the development costs and performance between \name and human Expert implementing attention operators on A100 GPU, which both utilize CuTe to achieve better performance.
Results demonstrate that \name not only reduces development time from months to minutes (a thousandfold reduction) but also achieves modest performance improvements.
% under the configuration of head dimension 64 and sequence length 1024. All developers directly utilize CUDA for operator development instead of relying on torch APIs, aiming to achieve better performance. Junior developer implements attention operator using CUDA Core, while senior developer leverages Tensor Core. The results demonstrate that \name reduces the development cost by $384\times$ (calculated based on an 8-hour workday) compared to senior developer, while achieving an $86\times$ performance improvement over junior developer.

% To validate the rationality and effectiveness of the designed Domain-Specific Language for Thinking, this study employs an ablation experiment to evaluate the DSL architecture. Specifically, we remove the DSL module, allowing the translator to directly generate the target CuTe code based on the input information, thereby assessing the accuracy of code generation in the absence of the DSL layer.
%为验证分层抽象设计在TL代码生成中的必要性，本研究设计并实施了系统的消融实验。具体而言，我们将原本多阶段的DSL生成过程简化为单阶段生成，即直接由大型语言模型（LLMs）输出TL代码，以此评估分层设计对系统性能的影响。实验结果表明，现有LLMs均无法通过单阶段生成完全正确的TL代码。基于系统性分析，我们在此重点列举两类典型错误，具体错误案例详见附录。
% \label{lb:TL}
\textbf{Impact on TL Designation}. To validate the necessity of hierarchical generation in \name design (first TL sketch then TL code), we design an ablation experiment to make LLMs directly output TL code. 
%and conducted a systematic ablation experiment. We turn the originally multi-stage process of TL generation into a single-stage process, where LLMs directly output TL code, thereby evaluating the impact of hierarchical design on system performance.
The results demonstrate that none of the existing LLMs is capable of generating entirely correct TL code in a single stage, and representative errors are like Reshape omission when generating fused computation and GEMM layout error.
% Through systematic analysis, there are two representative categories of errors, Reshape omission and GEMM error.
For more details refer to Appendix \ref{appendix:err}.

% Table \ref{tab:develop_ablation} presents a systematic comparative analysis of development efficiency and computational performance for FlashAttention implementation on A100 GPU (head dimension 64, sequence length 1024). The experimental groups include: 1) Claude 3.5 and DeepSeek-R1 with CoT; 2) Human experts with over two years of Tensor Core development experience; 3) \name. Key findings reveal that \name achieves 175.6 TFLOPS within 20 minutes, demonstrating 96× faster development speed(Working days calculated at 8 hours per day) and 4.7× higher performance compared to human experts. Baseline models using CoT show capability in generating CUDA code, but their computational efficiency remains below 5.4\% of the human expert baseline.

\begin{table}[]
\centering{
\resizebox{0.7\linewidth}{!}{
\begin{tabular}{l|rrr}
\toprule
Sequence Length     & \multicolumn{1}{r}{512} & \multicolumn{1}{r}{1k} & \multicolumn{1}{r}{2k} \\ \midrule
DeepSeek-V3     & 0.02   &     0.004     &   -    \\
+ CoT     &  0.12    &  0.27 &  0.52  \\
 + \name     &107.4      &   134.6    & 154.7    \\ \bottomrule
\end{tabular}}
}
\caption{CUDA implementation performance (TFLOPS) comparison of MHA with causal mask and head dimension 64 on A100 GPUs between CoT and our method.}
\label{tbl:cot}
\end{table}
\begin{table}[]
    \centering
    \resizebox{\linewidth}{!}{
    \begin{tabular}{l|rrrrrr}
\toprule
    Sequence Length & 512& 1k& 2k & 4k & 8k & 16k \\ \midrule
    Performance & 224.8 & 241.1 & 248.3 & 254.6 & 255.1 & 257.9 \\ \midrule
    \end{tabular}
    }
    \caption{Performance (TFLOPS) of MHA with causal mask on L40S GPU using \name under the configuration of head dimension 128, FP8 datatype.}
    \label{tab:fp8}
    \vspace{-10pt}
\end{table}
\subsection{Case Study}
% \textbf{Attention Operator Support for Heterogeneous Architectures and Datatypes.} 
% To demonstrate the generalizability of our approach, we  two specialized attention operator implementations targeting distinct application scenarios: 1) FP8-compatible MHA operator, and 2) T4 GPU-compatible attention operators. The experimental results are shown in Table \ref{tab:fp8} and Appendix \ref{sec:appendix}.
To demonstrate the generality of \name, we validate our approach under two extra scenarios, 1) FP8-compatible MHA operator (unsupported by other libraries), and 2) T4 GPU-compatible attention operators. Results in Table \ref{tab:fp8} and Table \ref{tbl:t4} in Appendix show that LLMs leveraging LLM-TL successfully generated high-performance attention operator implementations.
%To further validate the generality of the method, this study also extended the testing to variants of other attention operators. Specifically, we focused on evaluating the MLA operator, which has been widely adopted in the latest generation of LLMs (e.g., DeepSeek-V3) and has demonstrated significant advantages in reducing training and inference costs. To adapt to the characteristics of the MLA operator, we made targeted optimizations to the existing pipeline, primarily by introducing the Rotary Position Embedding (RoPE)~\cite{su2024roformer} module into the attention mechanism, to comprehensively assess its performance and output quality in code generation tasks.

\section{Conclusion}
In this paper, we propose a novel framework called \name to systematically unlocks LLMs' potential for high-performance attention operator generation.
% Through extensive experimental comparison, we find that the key challenge resides in enabling the LLM to fully comprehend the execution flow of both the algorithm and the hardware by TL-code, and adapting different GPU architectures by translating this TL-Code.
Extensive experiments identify that \name enables LLMs to fully understand algorithm-hardware execution flows via TL-code and adapt across GPU architectures through TL-code translation.
The results show that the code generated through \name could outperform hand-optimized libraries crafted by human experts while reducing months development time to minutes.
In summary, \name not only extends the capacities of LLMs on complex code optimization, but also achieves a self-optimizing paradigm for generating high-performance attention operators in attention-centric algorithms.
\section*{Limitations} 
Our proposed LLM-TL and its corresponding workflow enable LLMs to comprehend and generate high-performance attention operator implementations across various GPUs. Compared to vanilla LLMs, this approach demonstrates significant performance improvements, achieving results that are comparable to or surpass those of state-of-the-art handcrafted libraries. However, our method does have some limitations.

Due to resource constraints, although we conducted extensive experiments across multiple generations of GPUs, testing on the latest H100 was not performed. Additionally, while LLM-TL and its workflow are designed to be generalizable to complex operators on GPUs, we did not further evaluate its applicability to other types of operators, given that FlashAttention has been a highly complex and widely used operator already.

To address these limitations, our future work will focus on extending LLM-TL to GPUs with newer architectures and applying it to a broader range of complex operators.

\section{Ackowledgement}

This work is partially supported by the Strategic Priority Research Program of the Chinese Academy of Sciences (Grants No.XDB0660300, XDB0660301, XDB0660302), the NSF of China (Grants No.62525203, 92364202, U22A2028, 62302483), Major Program of lSCAS (Grant No. ISCAS.ZD-202402), CAS Project for Young Scientists in Basic Research (YSBR-029) and Youth Innovation Promotion Association CAS.

\section*{Ethics Statement}
This research on generating high-performance attention operators using LLMs considers ethical implications, including bias, misuse, and societal impact. We emphasize transparency, responsible data practices, and encourage ethical deployment of AI in software engineering.

% \section*{References}
\bibliography{acl_arxiv}

\newpage

\appendix

\section{Additional Evaluations}
% \section{Evaluations on Other Hardware Architectures}
\label{sec:appendix}

\begin{table*}[h]
\centering
\resizebox{\textwidth}{!}{

\begin{tabular}{c|l|rrrrrr|rrrrrr}
\toprule
 \multicolumn{2}{c|}{Head Dimension}        & \multicolumn{6}{c|}{64}                                                                             & \multicolumn{6}{c}{128}                                                                             \\ \midrule
                                  \multicolumn{2}{c|}{Sequence Length}      & 512            & 1k             & 2k             & 4k             & 8k             & 16k            & 512            & 1k             & 2k             & 4k             & 8k             & 16k            \\ \midrule

 \multirow{6}{*}{\begin{tabular}[c]{@{}c@{}}Masked \\ MHA \end{tabular}} & cuDNN             & 8.11    & 10.84  & 12.13  & 13.22   & 13.69  & 13.83    & 7.73    & 10.41  & 12.14  & 13.03  & 13.76   & 13.01   \\
                  & FlexAttention & \textbf{10.82}    & 13.45   & 16.31  & 18.52   & 19.84  & 20.47    & 7.81     & 10.53   & 12.62   & 14.26  & 15.02   & 14.91   \\
                     & flash-attn v1 & 8.68    & 9.85   & 12.81  & 12.81   & 13.83  & 13.25    & 7.8     & 8.77   & 9.88   & 10.68  & 10.68   & 10.53   \\
    & DeepSeek-V3             & 1.33    & 1.35   & 0.99   & 1.21    & OOM    & OOM      & 2.46    & 2.64   & 1.92   & 2.26    & 2.38     & OOM     \\
                     & \cellcolor[rgb]{0.925,0.925,0.925} DeepSeek-V3 + Ours            & \cellcolor[rgb]{0.925,0.925,0.925}{9.83} & \cellcolor[rgb]{0.925,0.925,0.925}\textbf{13.48} & \cellcolor[rgb]{0.925,0.925,0.925}\textbf{16.62} & \cellcolor[rgb]{0.925,0.925,0.925}\textbf{19.11} & \cellcolor[rgb]{0.925,0.925,0.925}\textbf{20.72} & \cellcolor[rgb]{0.925,0.925,0.925}\textbf{21.43} & \cellcolor[rgb]{0.925,0.925,0.925}\textbf{9.92} & \cellcolor[rgb]{0.925,0.925,0.925}\textbf{13.67} & \cellcolor[rgb]{0.925,0.925,0.925}\textbf{16.82} & \cellcolor[rgb]{0.925,0.925,0.925}\textbf{18.61} & \cellcolor[rgb]{0.925,0.925,0.925}\textbf{19.62} & \cellcolor[rgb]{0.925,0.925,0.925}\textbf{19.07} \\
                     & \cellcolor[rgb]{0.925,0.925,0.925}      & \cellcolor[rgb]{0.925,0.925,0.925}{$\uparrow$7.39×} & \cellcolor[rgb]{0.925,0.925,0.925}\textbf{$\uparrow$9.99×} & \cellcolor[rgb]{0.925,0.925,0.925}\textbf{$\uparrow$16.79×} & \cellcolor[rgb]{0.925,0.925,0.925}\textbf{$\uparrow$15.80×} & \cellcolor[rgb]{0.925,0.925,0.925}- & \cellcolor[rgb]{0.925,0.925,0.925}- & \cellcolor[rgb]{0.925,0.925,0.925}\textbf{$\uparrow$4.03×} & \cellcolor[rgb]{0.925,0.925,0.925}\textbf{$\uparrow$5.18×} & \cellcolor[rgb]{0.925,0.925,0.925}\textbf{$\uparrow$8.76×} & \cellcolor[rgb]{0.925,0.925,0.925}\textbf{$\uparrow$8.23×} & \cellcolor[rgb]{0.925,0.925,0.925}\textbf{$\uparrow$8.24×} & \cellcolor[rgb]{0.925,0.925,0.925}- \\
 \midrule
                \multirow{6}{*}{\begin{tabular}[c]{@{}c@{}}Masked\\ GQA\end{tabular}} 
               & cuDNN             & 8.21    & 10.42  & 12.13  & 13.03   & 13.5   & 13.85    & 7.74    & 10.41  & 11.63  & 12.84  & 13.49   & 12.93   \\
                     & FlexAttention & \textbf{10.51}    & 13.45   & 16.17  & 18.54   & 20.13  & 20.58    & 7.92     & 10.24   & 12.43   & 14.48  & 15.13   & 14.89   \\
                     & flash-attn v1 & 6.85    & 7.93   & 12.79  & 13.22   & 13.56  & 13.46    & 5.3     & 9.37   & 9.85   & 10.62  & 10.86   & 10.51   \\
              & DeepSeek-V3             & 1.05    & 1.06   & 0.82   & 0.97    & OOM    & OOM      & 1.97    & 2.05   & 1.58   & 1.81   & 1.93    & OOM     \\
        & \cellcolor[rgb]{0.925,0.925,0.925} DeepSeek-V3 + Ours              & \cellcolor[rgb]{0.925,0.925,0.925}{9.76} & \cellcolor[rgb]{0.925,0.925,0.925}\textbf{13.68} & \cellcolor[rgb]{0.925,0.925,0.925}\textbf{16.45} & \cellcolor[rgb]{0.925,0.925,0.925}\textbf{19.13} & \cellcolor[rgb]{0.925,0.925,0.925}\textbf{20.81} & \cellcolor[rgb]{0.925,0.925,0.925}\textbf{21.58} & \cellcolor[rgb]{0.925,0.925,0.925}\textbf{10.73} & \cellcolor[rgb]{0.925,0.925,0.925}\textbf{14.08} & \cellcolor[rgb]{0.925,0.925,0.925}\textbf{16.32} & \cellcolor[rgb]{0.925,0.925,0.925}\textbf{18.03} & \cellcolor[rgb]{0.925,0.925,0.925}\textbf{19.15} & \cellcolor[rgb]{0.925,0.925,0.925}\textbf{18.7} \\
        & \cellcolor[rgb]{0.925,0.925,0.925}         & \cellcolor[rgb]{0.925,0.925,0.925}{$\uparrow$9.30×} & \cellcolor[rgb]{0.925,0.925,0.925}\textbf{$\uparrow$12.91×} & \cellcolor[rgb]{0.925,0.925,0.925}\textbf{$\uparrow$20.06×} & \cellcolor[rgb]{0.925,0.925,0.925}\textbf{$\uparrow$19.72×} & \cellcolor[rgb]{0.925,0.925,0.925}- & \cellcolor[rgb]{0.925,0.925,0.925}- & \cellcolor[rgb]{0.925,0.925,0.925}\textbf{$\uparrow$5.45×} & \cellcolor[rgb]{0.925,0.925,0.925}\textbf{$\uparrow$6.87×} & \cellcolor[rgb]{0.925,0.925,0.925}\textbf{$\uparrow$10.33×} & \cellcolor[rgb]{0.925,0.925,0.925}\textbf{$\uparrow$9.96×} & \cellcolor[rgb]{0.925,0.925,0.925}\textbf{$\uparrow$9.92×} & \cellcolor[rgb]{0.925,0.925,0.925}- \\
 \midrule
   \multirow{6}{*}{\begin{tabular}[c]{@{}c@{}}Masked\\    MQA\end{tabular}} & cuDNN             & 8.41    & 10.45  & 11.77  & 12.95   & 13.43  & 13.78    & 8.39    & 10.32  & 11.62  & 13.03  & 13.6    & 12.82   \\
                     & FlexAttention & \textbf{10.67}    & 13.42  & 16.44  & 18.39   & 19.91  & 20.27    & 7.89    & 10.42   & 12.64   & 14.12  & 14.98   & 14.95   \\
                     & flash-attn v1 & 5.73    & 11.53  & 12.65  & 13.74   & 13.59  & 13.38    & 5.27    & 9.42   & 9.95   & 10.71  & 10.32   & 10.47   \\
              & DeepSeek-V3             & 1.35    & 1.36   & 0.99   & 1.21    & OOM    & OOM      & 2.52    & 2.61   & 1.91   & 2.26   & 2.37    & OOM     \\
                     & \cellcolor[rgb]{0.925,0.925,0.925}  DeepSeek-V3 + Ours             & \cellcolor[rgb]{0.925,0.925,0.925}{9.65} & \cellcolor[rgb]{0.925,0.925,0.925}\textbf{13.61} & \cellcolor[rgb]{0.925,0.925,0.925}\textbf{17.11} & \cellcolor[rgb]{0.925,0.925,0.925}\textbf{18.84} & \cellcolor[rgb]{0.925,0.925,0.925}\textbf{20.59} & \cellcolor[rgb]{0.925,0.925,0.925}\textbf{21.27} & \cellcolor[rgb]{0.925,0.925,0.925}\textbf{10.19} & \cellcolor[rgb]{0.925,0.925,0.925}\textbf{14.01} & \cellcolor[rgb]{0.925,0.925,0.925}\textbf{16.7} & \cellcolor[rgb]{0.925,0.925,0.925}\textbf{17.92} & \cellcolor[rgb]{0.925,0.925,0.925}\textbf{19.17} & \cellcolor[rgb]{0.925,0.925,0.925}\textbf{18.79} \\
                      & \cellcolor[rgb]{0.925,0.925,0.925}      & \cellcolor[rgb]{0.925,0.925,0.925}{$\uparrow$7.15×} & \cellcolor[rgb]{0.925,0.925,0.925}\textbf{$\uparrow$10.00×} & \cellcolor[rgb]{0.925,0.925,0.925}\textbf{$\uparrow$17.28×} & \cellcolor[rgb]{0.925,0.925,0.925}\textbf{$\uparrow$15.57×} & \cellcolor[rgb]{0.925,0.925,0.925}- & \cellcolor[rgb]{0.925,0.925,0.925}- & \cellcolor[rgb]{0.925,0.925,0.925}\textbf{$\uparrow$4.04×} & \cellcolor[rgb]{0.925,0.925,0.925}\textbf{$\uparrow$5.37×} & \cellcolor[rgb]{0.925,0.925,0.925}\textbf{$\uparrow$8.74×} & \cellcolor[rgb]{0.925,0.925,0.925}\textbf{$\uparrow$7.93×} & \cellcolor[rgb]{0.925,0.925,0.925}\textbf{$\uparrow$8.09×} & \cellcolor[rgb]{0.925,0.925,0.925}- \\

  \midrule
  \midrule
 \multirow{6}{*}{MHA} & cuDNN             & 10.47   & 12.51  & 13.01  & 13.61   & 13.95  & 13.76    & 10.81   & 12.35  & 12.95  & 13.51  & 13.53   & 13.58   \\
      & FlexAttention & \textbf{14.42}   & \textbf{16.91}  & \textbf{18.77}  & \textbf{19.85}   & \textbf{20.13}  & \textbf{20.25}    & 11.34    & 13.50  & 14.83  & 15.67  & 15.74   & 15.42   \\

     & flash-attn v1 & 12.25   & 15.27  & 14.28  & 14.25   & 14.25  & 14.46    & 8.78    & 11.03  & 10.95  & 10.61  & 10.38   & 10.78   \\
      & DeepSeek-V3             & 4.23    & 4.54   & 3.18   & 4.07    & OOM   & OOM      & 7.87    & 8.23   & 6.11   & 7.16   & 7.48     & OOM     \\
      & \cellcolor[rgb]{0.925,0.925,0.925}  DeepSeek-V3 + Ours            & \cellcolor[rgb]{0.925,0.925,0.925}{13.35} & \cellcolor[rgb]{0.925,0.925,0.925}{15.68} & \cellcolor[rgb]{0.925,0.925,0.925}{18.35} & \cellcolor[rgb]{0.925,0.925,0.925}{19.36} & \cellcolor[rgb]{0.925,0.925,0.925}{19.59} & \cellcolor[rgb]{0.925,0.925,0.925}{19.67} & \cellcolor[rgb]{0.925,0.925,0.925}\textbf{15.8} & \cellcolor[rgb]{0.925,0.925,0.925}\textbf{17.79} & \cellcolor[rgb]{0.925,0.925,0.925}\textbf{18.59} & \cellcolor[rgb]{0.925,0.925,0.925}\textbf{18.68} & \cellcolor[rgb]{0.925,0.925,0.925}\textbf{18.86} & \cellcolor[rgb]{0.925,0.925,0.925}\textbf{19.13} \\
    & \cellcolor[rgb]{0.925,0.925,0.925}       & \cellcolor[rgb]{0.925,0.925,0.925}{$\uparrow$3.16×} & \cellcolor[rgb]{0.925,0.925,0.925}{$\uparrow$3.45×} & \cellcolor[rgb]{0.925,0.925,0.925}{$\uparrow$5.77×} & \cellcolor[rgb]{0.925,0.925,0.925}{$\uparrow$4.76×} & \cellcolor[rgb]{0.925,0.925,0.925}- & \cellcolor[rgb]{0.925,0.925,0.925}- & \cellcolor[rgb]{0.925,0.925,0.925}\textbf{$\uparrow$2.01×} & \cellcolor[rgb]{0.925,0.925,0.925}\textbf{$\uparrow$2.16×} & \cellcolor[rgb]{0.925,0.925,0.925}\textbf{$\uparrow$3.04×} & \cellcolor[rgb]{0.925,0.925,0.925}\textbf{$\uparrow$2.61×} & \cellcolor[rgb]{0.925,0.925,0.925}\textbf{$\uparrow$2.52×} & \cellcolor[rgb]{0.925,0.925,0.925}- \\
 \midrule
 \multirow{6}{*}{GQA} 
& cuDNN             & 10.63   & 12.35  & 13.14  & 13.39   & 13.67  & 13.52    & 10.97   & 12.28  & 13.02  & 13.52  & 13.64   & 13.47   \\
     & FlexAttention & \textbf{14.43}    & 16.72  & \textbf{18.87}  & \textbf{20.12}   & \textbf{20.34} & \textbf{20.24} & 11.43    & 13.74   & 14.95  & 15.52  & 15.64  & 15.51    \\

      & flash-attn v1 & 8.89    & 15.43  & 14.41  & 14.48   & 14.35  & 14.52    & 10.98   & 11.01  & 10.95  & 11.01  & 10.33   & 10.8    \\
      & DeepSeek-V3             & 2.47    & 2.49   & 2.08   &{OOM} & OOM    & OOM      & 4.56    & 4.77   & 3.97   & 4.48   &{OOM}  & OOM     \\
      & \cellcolor[rgb]{0.925,0.925,0.925}  DeepSeek-V3 + Ours              & \cellcolor[rgb]{0.925,0.925,0.925}{13.82} & \cellcolor[rgb]{0.925,0.925,0.925}\textbf{16.79} & \cellcolor[rgb]{0.925,0.925,0.925}{18.47} & \cellcolor[rgb]{0.925,0.925,0.925}{19.68} & \cellcolor[rgb]{0.925,0.925,0.925}{19.51} & \cellcolor[rgb]{0.925,0.925,0.925}{19.47} & \cellcolor[rgb]{0.925,0.925,0.925}\textbf{15.81} & \cellcolor[rgb]{0.925,0.925,0.925}\textbf{17.59} & \cellcolor[rgb]{0.925,0.925,0.925}\textbf{18.82} & \cellcolor[rgb]{0.925,0.925,0.925}\textbf{19.6} & \cellcolor[rgb]{0.925,0.925,0.925}\textbf{19.53} & \cellcolor[rgb]{0.925,0.925,0.925}\textbf{19.58} \\
      & \cellcolor[rgb]{0.925,0.925,0.925}         & \cellcolor[rgb]{0.925,0.925,0.925}{$\uparrow$5.60×} & \cellcolor[rgb]{0.925,0.925,0.925}\textbf{$\uparrow$6.74×} & \cellcolor[rgb]{0.925,0.925,0.925}{$\uparrow$8.88×} & \cellcolor[rgb]{0.925,0.925,0.925}- & \cellcolor[rgb]{0.925,0.925,0.925}- & \cellcolor[rgb]{0.925,0.925,0.925}- & \cellcolor[rgb]{0.925,0.925,0.925}\textbf{$\uparrow$3.47×} & \cellcolor[rgb]{0.925,0.925,0.925}\textbf{$\uparrow$3.69×} & \cellcolor[rgb]{0.925,0.925,0.925}\textbf{$\uparrow$4.74×} & \cellcolor[rgb]{0.925,0.925,0.925}\textbf{$\uparrow$4.38×} & \cellcolor[rgb]{0.925,0.925,0.925}- & \cellcolor[rgb]{0.925,0.925,0.925}- \\
 \midrule 
 \multirow{6}{*}{MQA} 
& cuDNN             & 10.79   & 12.17  & 13.23  & 13.62   & 13.85  & 13.69    & 10.45   & 12.39  & 13.03  & 13.25  & 13.48   & 13.38   \\
         & FlexAttention & \textbf{14.61}   & \textbf{16.94}  & 18.57  & \textbf{19.93}   & \textbf{20.31}  & \textbf{20.14}    & 11.42   & 13.85  & 14.64  & 15.28  & 15.76   & 15.56   \\
      & flash-attn v1 & 11.23   & 15.32  & 14.59  & 14.12   & 14.14  & 14.51    & 11.19   & 11.06  & 11.01  & 11.02  & 10.72   & 10.88   \\
         & DeepSeek-V3             & 4.33    & 4.41   & 3.15   & 3.97    & OOM    & OOM      & 7.91    & 8.31   & 5.99   & 7.11   & 7.32    & OOM     \\
          & \cellcolor[rgb]{0.925,0.925,0.925} DeepSeek-V3 + Ours             & \cellcolor[rgb]{0.925,0.925,0.925}{13.58} & \cellcolor[rgb]{0.925,0.925,0.925}{16.34} & \cellcolor[rgb]{0.925,0.925,0.925}\textbf{18.77} & \cellcolor[rgb]{0.925,0.925,0.925}{19.67} & \cellcolor[rgb]{0.925,0.925,0.925}{19.79} & \cellcolor[rgb]{0.925,0.925,0.925}{19.62} & \cellcolor[rgb]{0.925,0.925,0.925}\textbf{14.94} & \cellcolor[rgb]{0.925,0.925,0.925}\textbf{16.39} & \cellcolor[rgb]{0.925,0.925,0.925}\textbf{18.14} & \cellcolor[rgb]{0.925,0.925,0.925}\textbf{19.4} & \cellcolor[rgb]{0.925,0.925,0.925}\textbf{19.08} & \cellcolor[rgb]{0.925,0.925,0.925}\textbf{19.46} \\
          & \cellcolor[rgb]{0.925,0.925,0.925}      & \cellcolor[rgb]{0.925,0.925,0.925}{$\uparrow$3.14×} & \cellcolor[rgb]{0.925,0.925,0.925}{$\uparrow$3.71×} & \cellcolor[rgb]{0.925,0.925,0.925}\textbf{$\uparrow$5.96×} & \cellcolor[rgb]{0.925,0.925,0.925}{$\uparrow$4.95×} & \cellcolor[rgb]{0.925,0.925,0.925}- & \cellcolor[rgb]{0.925,0.925,0.925}- & \cellcolor[rgb]{0.925,0.925,0.925}\textbf{$\uparrow$1.89×} & \cellcolor[rgb]{0.925,0.925,0.925}\textbf{$\uparrow$1.97×} & \cellcolor[rgb]{0.925,0.925,0.925}\textbf{$\uparrow$3.03×} & \cellcolor[rgb]{0.925,0.925,0.925}\textbf{$\uparrow$2.73×} & \cellcolor[rgb]{0.925,0.925,0.925}\textbf{$\uparrow$2.61×} & \cellcolor[rgb]{0.925,0.925,0.925}- \\
               \bottomrule
\end{tabular}
}
\caption{
Performance (TFLOPS)  comparison across different dimensions, attention operators and the presence or absence of causal masks on T4 GPU.
}
\label{tbl:t4}
\end{table*}

\begin{table*}[h]
\centering
\resizebox{0.7\textwidth}{!}{
\begin{tabular}{c|rrrrrr}
\toprule
                                  \multicolumn{1}{c|}{Sequence Length}      & 512            & 1k             & 2k             & 4k             & 8k             & 16k             \\ \midrule
% \multirow{6}{*}{\begin{tabular}[c]{@{}c@{}}Llama2 7B \\ (32 Q-heads/32 KV-heads)\end{tabular}} 
\multicolumn{7}{c}{Llama2 7B (32 Q-heads/32 KV-heads/128 Head-dimension) } \\
\cmidrule{1-7} 

cuDNN             & 112.4   & 142.6  & 164.1  & 176.8   & 197.2  & 201.7     \\
FlexAttention     & 82.8   & 108.6  & 128.6  & 150.2   & 158.4  & 167.1   \\
flash-attn v2     & 122.5   & 152.5  & 173.4  & 186.3   & 201.5  & 207.3   \\
DeepSeek-V3      & 14.6    & 15.1   & 10.9   & 13.3    & 14.6   & 15.1       \\
\rowcolor[rgb]{0.925,0.925,0.925} DeepSeek-V3 + Ours             & \textbf{137.1}          & \textbf{160.6}         & \textbf{180.3}         & \textbf{186.7}          & 198.3         & 202.7            \\ 
\rowcolor[rgb]{0.925,0.925,0.925}                    & \textbf{$\uparrow$9.39×} & \textbf{$\uparrow$10.64×} & \textbf{$\uparrow$16.54×} & \textbf{$\uparrow$14.04×} & $\uparrow$13.58× 
  & {$\uparrow$13.42×}  \\
\cmidrule{1-7} 
% \multirow{6}{*}{\begin{tabular}[c]{@{}c@{}}Qwen2.5 72B \\ (64 Q-heads/8 KV-heads) \end{tabular}}
\multicolumn{7}{c}{Qwen2.5 72B (64 Q-heads/8 KV-heads/128 Head-dimension) } \\
\cmidrule{1-7} 

cuDNN             & 116.3   & 148.3  & 176.1  & 191.8   & 202.1  & 207.9     \\
FlexAttention     & 86.1   & 117.4  & 136.7  & 154.7   & 168.1  & 172.4   \\
flash-attn v2     & 127.1   & 157.8  & 181.5  & \textbf{201.8}   & \textbf{216.0}  & \textbf{222.5}   \\
DeepSeek-V3       & 10.9    & 11.1   & 8.8   & 10.2    & 11.2   & OOM       \\
\rowcolor[rgb]{0.925,0.925,0.925} DeepSeek-V3 + Ours             &\textbf {140.2}          & \textbf{165.1}         & \textbf{186.4}         & 192.9          & 201.8         & 205.1            \\ 
\rowcolor[rgb]{0.925,0.925,0.925}           & \textbf{$\uparrow$12.86×} & \textbf{$\uparrow$14.87×} & \textbf{$\uparrow$21.18×} & $\uparrow$18.91× & $\uparrow$18.02× 
  & -   \\
\cmidrule{1-7} 
% \multirow{6}{*}{\begin{tabular} [c]{@{}c@{}} Llama3.1 405B \\ (128 Q-heads/8 KV-heads) \end{tabular}}
\multicolumn{7}{c}{Llama3.1 405B (128 Q-heads/8 KV-heads/128 Head-dimension) } \\
\cmidrule{1-7} 
cuDNN             & 121.9   & 157.2  & 183.4  & 196.7   & 206.4  & 211.2     \\
FlexAttention     & 93.2   & 117.2  & 144.4  & 157.7   & 169.7  & 175.3   \\
flash-attn v2     & 129.1   & 164.1  & \textbf{192.2}  & \textbf{209.1}   & \textbf{221.1}  & \textbf{225.3}   \\
DeepSeek-V3       & 11.1    & 11.2   & 8.9   & 10.2    & OOM   & OOM       \\
\rowcolor[rgb]{0.925,0.925,0.925} DeepSeek-V3 + Ours            & \textbf{146.5}          & \textbf{168.4}         & 187.5         & 200.3          & 204.4         & 206.9            \\ 
\rowcolor[rgb]{0.925,0.925,0.925}                 & \textbf{$\uparrow$13.20×} & \textbf{$\uparrow$15.03×} & $\uparrow$21.07× & $\uparrow$19.64× & - 
  & -   \\

                \midrule
\end{tabular}
}

\caption{Performance (TFLOPS) compared with handcraft libraries in different configuration on A100 GPU. The head dimension for all three LLMs is consistently set to 128 and all of attention operators are applied with causal mask. 
%Configuration: Llama2 7B (32 query heads, key and value heads), Qwen2.5 72B (64 query heads/8 key and value heads) and Llama3.1 405B (128 query heads/8 key and value heads)}
}
\label{tbl:llms}
\end{table*}

\begin{table}[htbp]
\resizebox{\linewidth}{!}{
\begin{tabular}{l|rrrrrr}
\toprule
Sequence Length    & 512   & 1k    & 2k    & 4k    & 8k    & 16k   \\ \midrule

Naive NSA & 0.84  & 1.68  & 3.35  & 6.61  & 13.34  & 26.29  \\ 
\rowcolor[rgb]{0.925,0.925,0.925} ours     & \textbf { 0.67}  & \textbf{1.26}  & \textbf{2.59} & \textbf{5.25}  & \textbf{10.59}  & \textbf{21.27} \\ 
            \rowcolor[rgb]{0.925,0.925,0.925} & \textbf{$\uparrow$1.25×} & \textbf{$\uparrow$1.33×} & \textbf{$\uparrow$1.29×} & \textbf{$\uparrow$1.26×} & \textbf{$\uparrow$1.26×} & \textbf{$\uparrow$1.24×}
  \\ 
  \cmidrule{1-7}
\end{tabular}
}
\caption{The NSA evaluations of latency(s) on A100 with head dimension of 128. We compared with naive pyTorch implementation of NSA.}
% \caption{MLA with causal mask performance (TFLOPS) compared with handcraft libraries on A100 GPU under the configuration of head dimension 128, RoPE dimension 64 with causal mask.}
\label{tbl:nsa}
\end{table}

We further conduct extensive experiments across various NVIDIA hardware architectures to validate the robustness and scalability of our proposed method. Comparing our automatically generated code with three implementations on T4 GPU. 
The experimental results, as presented in Table \ref{tbl:t4}, demonstrate that the code generated by our method consistently achieves superior performance metrics compared to all four implementations. This performance advantage is maintained across all evaluated dimensions, indicating the robustness and effectiveness of our approach on the NVIDIA T4 GPU architecture. 

Moreover, we extend our experiments to assess the applicability of our method on alternative attention mechanisms, NSA\cite{yuan2025native}. The results, summarized in Table \ref{tbl:nsa}, show that our approach maintains its performance advantage when applied to these attention variants. For NSA, the automatically generated code consistently outperforms manually optimized baselines in terms of latency and throughput, reaffirming the versatility and adaptability of our method across diverse attention architectures.

% \subsection{Evaluations on Other Attention Variants}

\section{Ablation on TL Code Generation stage}
We conduct an ablation study to validate the necessity of the hierarchical generation in TL Code Generation Stage (i.e., first generate TL sketch then TL code).
Results demonstrate that existing LLMs fails to directly generate TL code.
\label{appendix:err}
Listing \ref{lst:omission} and Listing \ref{lst:GEMM} provide the failure cases.

\textbf{Reshape omission.} LLMs occasionally fail to perform the necessary layout reshape operation on the result matrix generated by the first GEMM operator before proceeding to the second GEMM operator. Although the output matrix of the first GEMM operator and one of the input matrices of the second GEMM operator are mathematically consistent, their memory layouts differ significantly due to the use of MMA instructions. Therefore, a layout reshape operation must be introduced to reorganize the matrix layout to meet the requirements of MMA instructions. The lack of TL Sketch prevents LLMs from recognizing the critical layout transformation, causing reshape omissions and computation errors.

\textbf{GEMM error.} LLMs fail to rigorously distinguish between TL symbolic representations and their corresponding physical memory layout mappings. Specifically, although the first GEMM operator in the TL layer is defined as $Q$ multiplied by $K^T$, the matrix $K$ physically retains its original layout in memory due to the characteristics of the MMA instruction set. Nevertheless, the TL abstraction layer must maintain formal transpose notation to guide subsequent translation processes properly. Due to the absence of TL Sketch intermediate representation generation, LLMs fail to effectively bridge the semantic gap between algorithmic abstraction and hardware implementation layers. This results in the conflation of formal transpose notation in TL with actual physical memory layouts (without explicit transposition), leading to the loss of critical semantic constraints during TL-to-CuTe translation and ultimately causing implementation-level mismatches.

\begin{lstlisting}[caption=A Case of Reshape Omission., label=lst:omission]
Compute GEMM Q_shared, K_shared.T and get S
if i < (kv_len/BN) - 1
  Copy K (BN, HeadDim) in coordinate [L = i+1] from global to shared
end
Compute Softmax S
// No reshape!
Compute GEMM S, V_shared and accumulate O_register
\end{lstlisting}
\begin{lstlisting}[caption=A Case of GEMM Error., label=lst:GEMM]
Compute GEMM Q_shared, K_shared and get S //K_shared should be transposed
Compute Softmax S with Smax and Ssum
Reshape S from (MMA_C, MMA_M, MMA_N) to (MMA_A, MMA_M, MMA_N_new)
Compute GEMM S, V_shared and accumulate O_reg
\end{lstlisting}

\section{Real-World LLM Configuration}
\label{appendix:llm}
Currently, LLMs predominantly employ MHA and GQA mechanisms in their transformer backbone. While these models uniformly adopt a head dimension of 128 across different architectures, significant variations exist in their configuration of query heads versus key and value heads. To systematically evaluate \name on these architectural distinctions, we select three representative models spanning diverse scales and configurations: Llama2 7B (32 query heads, key and value heads), Qwen2.5 72B (64 query heads / 8 key and value heads), and Llama3.1 405B (128 query heads / 8 key and value heads).The experimental results, as presented in Table \ref{tbl:llms}, demonstrate that the code generated by \name consistently achieves superior performance metrics compared to all four implementations. This performance advantage is maintained across all evaluated configuration, indicating the robustness and effectiveness of \name on different heads. 

\section{Prompts}

Some prompts for guiding TL Code generating and reasoning are shown in Listing \ref{lst:TLgen} and Listing \ref{lst:TLreason}, respectively.

For TL Code generating, take \textit{Copy} statement as an example, we first instruct LLMs about the memory hierarchies of GPUs, to establish a foundational understanding of data movement across these hierarchies. Additionally, we introduce the architecture and operational principles of Tensor Core, emphasizing their role in MMA operations through warp-level parallelism and efficient register utilization. Following this, we provide the precise syntax and structure of TL Code, enabling LLMs to generate hardware-aware implementations correctly and effectively.

For TL Code reasoning, we instruct LLMs to derive specific dimensional information to refine the execution process of attention operator. Take \textit{Copy} statement again as an example, we first introduce parameter-related variable allocation mechanisms. Then, we let LLMs perform different operations based on the different cases of the \textit{Copy} statements, while the specific details such as dimensional information are generated by the LLMs themselves.

\begin{lstlisting}[caption={Prompt for guiding LLMs to generate TL Code.}, label=lst:TLgen]
Use the following basic TL statements to describe the provided algorithm flow. Focus solely on describing the hardware execution process of the algorithm on the GPU, without adding excessive complex information.
Here are two basic statements of the TL, `Copy` and `Compute`:
### Copy
The term "Copy" is used to denote the transfer of data between different levels of storage hierarchy in hardware. In GPUs, there are three primary storage levels:
- **global memory:** Global memory is high-capacity, high-latency video memory used for storing data shared by all threads, typically for holding complete matrices or large-scale datasets.
- **shared memory:** Shared memory is high-speed, low-latency on-chip memory used for data sharing within a thread block. It is typically used to store a small portion of data required for collaborative computation by the current thread block, loaded from global memory.
- **register:** Registers are the fastest private storage units, used for storing thread-local variables and temporary data. In CUDA Tensor Core operations, registers are directly involved in matrix multiply-accumulate (MMA) computations, where each thread fetches its assigned data from shared memory and stores it in registers.
In this context, a "Copy" operation requires specifying the variable name as well as the source and destination addresses of the variable.  Here is the usage of `Copy` statement: 
```
Copy A from global to shared
```
The clause means load a block of the matrix `A` to the corresponding shared memory storage. 
### Compute
The term "compute" is used to represent computations performed on hardware. Computational descriptions include various types of operations, primarily arithmetic operations (addition, subtraction, multiplication, division), matrix multiplication (GEMM), accumulation, and others. Here are some typical computations:
- **GEMM:** Use *GEMM* (General Matrix Multiplication) to represent the multiplication operation of two matrices at the register storage level, leveraging the fast access characteristics of registers to achieve high-performance matrix computations. Use GEMM. This primitive can be used in the following manner:
    ```
    Compute GEMM A, B and get S
    Compute GEMM A, B and accumulate S
    ```
    Here we compute the GEMM result of A and B, and store the result to the variable S.
- **Regular computation:** We need some regular computation like **the four basic arithmetic operations**, and we can use these basic operations like this:
    ```
    Compute Multiply A, x and get new A
    Compute Multiply A, x and get B
    ```
    Here we use these clauses to define the **multiplication** operation, and the first means store the result back to A while the other means store to a new variable B.
- **Other operators:** Sometimes the users will define some other custom operators like softmax, and we can use it like:
    ```
    Compute Softmax A
    ```
In this context, "compute" first requires specifying the exact type of computation, along with the variables involved in the computation and the variable name for the result of the computation.
### Loop
When describing the execution flow of an algorithm, it is often necessary to use **for loops** to represent iterative operations of operators. In such cases, a **For statement** is used to describe these loops. The syntax `for i = 0:N ... end` is employed to indicate a loop that iterates **N times**, and **indentation** is used to control which code blocks are executed within the loop.
```
for i=0:N
	...
end
```
Now, based on the algorithm workflow provided by the user, analyze the memory access and computational behavior of the algorithm on the GPU, and use the aforementioned statements to represent the semantically enriched execution flow of the algorithm.
\end{lstlisting}

\begin{lstlisting}[caption={Prompt for guiding LLMs to analyse parameters in the TL Code and reason the TL Code.}, label=lst:TLreason]
Several fundamental statement types have been defined to represent the algorithm execution flow. Building upon these primitive constructs, we need to derive specific dimensional information to refine the algorithm execution process. Besides, to facilitate the generation of functionally correct final code, this step introduces parameter-related variable allocation mechanisms. We achieve this process by incorporating allocate statements, thereby ensuring proper allocation and management of memory resources. Specifically, the dimensional information to be derived includes:
### Copy
For the `Copy` statement, it is necessary to complete its parameter configuration based on the execution characteristics of algorithm on the GPU architecture. The optimization primarily focuses on access locations within global memory. 
- Global to shared
Prior to performing the `Copy` statement from or to global memory, it is necessary to fully characterize the matrix information in global memory. The Allocate statement can be used to represent the storage layout and attributes of the entire matrix in global memory like the follow:
```
Allocate A in global (M, K) with offset batch_offset 
Copy A from global to shared
...
```
For  matrix `A` with shape `(batch, M, N)`, here we allocate tensor `A` whose shape is (M, K). Assuming that each block will be responsible for a batch, the global that should be loaded will have an offset of batch_offset. It can be used in the final implementation code. This Allocate statement needs to be applied to every `Copy` statement involving global memory, unless the corresponding memory allocation has already been explicitly declared in the preceeding context.
For `Copy` statement itself, for instance, this clause:
```
Copy A from global to shared
```
If it is required to load a data block of size (BM, BK) from matrix `A` located at position L = i, the parameter configuration can be added in the following manner:
```
Copy A (BM, BK) in coordinate [L = i] from global to shared
```
The first clause represents the original Sketch description. By incorporating the aforementioned parameter information, the following complete implementation can be derived the second one.
Here you should note that: L = i represents the i -th block after tiling. For example, if A is a matrix with shape (M, BK),  `Copy` A (BM, BK) in coordinate [L = i] selects the i-th block, which is with the shape (BM, BK) . And there will be M / BM blocks in total.
### Compute
The dimensionality of the compute statement is directly determined by the dimensions of its input and output registers, thus requiring no additional dimensional information. However, there are still two aspects related to the compute statement that require further supplementation:
- **Declare intermediate variables:**  ****During the computation process, certain intermediate register variables serving as outputs require additional definition of their shape and storage attributes. Here is an example of GEMM:
    ```
    ...
    for i in 0:K:
    	Compute GEMM A, B and get C
    	...
    ```
    where it didn't declare what C is, so you should add a allocate clause before the for loop to represent the shape information of S in register.
    ```
    ...
    Allocate C in register (BM, BN)
    for i in 0:K:
    	Compute GEMM A, B and get C
    	...
    ```
    Here we allocate the C in register with the information of whole dimension of block memory, and the real register shape can be inferred from the whole size.
- **Fuse two GEMM(Add reshape):** In the scenario of fusing two consecutive GEMM operations, where the output of the first operation directly serves as the input to the second, it is necessary to transform the output of the first operation according to the specific computational scale to meet the input requirements of the second GEMM operation.
    Here is some prior knowledge: The layout of Tensor Core can be represented as (MMA, MMA_M, MMA_N). Here, MMA denotes the computational scale required for matrices A, B, and C in the GEMM operation, while MMA_M and MMA_N represent the repetition counts of the computation along the M and N dimensions, respectively.
    Here is an example of fuse:
    ```
    Compute GEMM E, F and get G
    ... //There might be other operations, such as `Compute Softmax G`
    Compute GEMM G, H and accumulate I
    ```
    
    Here `G` is the output of first GEMM and then used in second GEMM. There might be other operations between two GEMMs, such as softmax. 
    HOWEVER, whenever there is an association between two GEMMs, a reshape statement must be added. So you have to add a reshape statement according to layout G. Add the statement before GEMM-II, just like the example below:
    ```
    Compute GEMM E, F and get G
    ... //There might be other operations, such as `Compute Softmax G`
    Reshape G from (MMA_C, MMA_M, MMA_N) to (MMA_A, MMA_M, MMA_N_new) 
    Compute GEMM G, H and accumulate I
    ```
    In this example, the `G` of GEMM-I is one of the input of GEMM-II. Since C is the `A` matrix of tensor core, the MMA shape of G should be changed from `MMA_C` to `MMA_A` , and change the `MMA_N` to adapt this difference.
\end{lstlisting}

\end{document}